\documentclass{article} 
\usepackage{iclr2027_conference,times}

\usepackage{amsmath,amsfonts,bm}

\def\eqref#1{equation~\ref{#1}}

\def\1{\bm{1}}

\DeclareMathAlphabet{\mathsfit}{\encodingdefault}{\sfdefault}{m}{sl}
\SetMathAlphabet{\mathsfit}{bold}{\encodingdefault}{\sfdefault}{bx}{n}

\usepackage{hyperref}
\usepackage{url}
\usepackage{amsmath}
\usepackage{amssymb}
\usepackage{bm}
\usepackage{graphicx}
\usepackage{booktabs}
\usepackage[capitalize,noabbrev]{cleveref}
\usepackage{natbib}
\usepackage{multirow}
\usepackage[table]{xcolor}

\usepackage{colortbl}

\usepackage{enumitem}
\usepackage[normalem]{ulem}
\usepackage{xcolor}
\usepackage{wrapfig}

\crefname{equation}{Eq.}{Eqs.}
\Crefname{equation}{Eq.}{Eqs.}
\crefname{subsubsection}{Section}{Sections}
\Crefname{subsubsection}{Section}{Sections}
\crefname{subsubappendix}{Appendix}{Appendices}
\Crefname{subsubappendix}{Appendix}{Appendices}
\crefname{figure}{Fig.}{Figs.}
\Crefname{figure}{Fig.}{Figs.}
\crefname{table}{Table}{Tables}
\Crefname{table}{Table}{Tables}
\crefname{section}{Section}{Sections}
\Crefname{section}{Section}{Sections}
\crefname{subsection}{Section}{Sections}
\Crefname{subsection}{Section}{Sections}
\crefname{appendix}{Appendix}{Appendices}
\Crefname{appendix}{Appendix}{Appendices}
\crefname{subappendix}{Appendix}{Appendices}
\Crefname{subappendix}{Appendix}{Appendices}
\crefname{algorithm}{Algorithm}{Algorithms}
\Crefname{algorithm}{Algorithm}{Algorithms}

\definecolor{linkblue}{HTML}{0B5394}  
\definecolor{citegreen}{HTML}{1B7F5C}  
\definecolor{urlteal}{HTML}{2A6F8F}    
\hypersetup{
    colorlinks=true,
    pdfborder={0 0 0},
    linkcolor=linkblue,
    citecolor=citegreen,
    urlcolor=urlteal
}

\title{From Static to Dynamic: On-Policy Distillation from Image to Video Diffusion Models}

\author{
Bingqing Jiang$^{1}$\thanks{These authors contributed equally to this work.}
\quad
Li Luo$^{1}$\footnotemark[1]
\quad
Zichao Yu$^{1}$\footnotemark[1]
\quad
Yujin Han$^{1}$\footnotemark[1]
\quad
Zhaolong Su$^{2}$
\quad
Difan Zou$^{1}$\thanks{Corresponding author: dzou@cs.hku.hk}
\\[0.2em]
$^{1}$The University of Hong Kong
\\
$^{2}$Cornell University
}

\iclrfinalcopy 
\begin{document}

\maketitle
\fancyhead{}
\lhead{Preprint.}

\begin{abstract}

On-policy distillation (OPD) specializes pretrained video diffusion models through teacher supervision along the student's own generation trajectory.
Although large video models are natural teachers, developing specialized video experts can require costly video data and training, while querying them incurs substantially higher latency than querying image experts.
More readily available and cheaper to query, image experts offer a cost-effective alternative, particularly for largely temporal-agnostic capabilities such as aesthetics and OCR that admit frame-level supervision.
However, heterogeneous image and video latent spaces prevent direct supervision of intermediate student states, while image experts lack cross-frame motion supervision, making temporal consistency vulnerable to frame-level improvements.
In this paper, we propose \textbf{MILD}, a \textbf{M}otion-Preserving \textbf{I}mage-to-Video \textbf{L}atent \textbf{D}istillation framework that transfers specialized image expertise while preserving pretrained video dynamics.
MILD uses a learnable linear connector that aligns student latent states and predicted updates with those of image experts, enabling supervision transfer across heterogeneous latent spaces.
We further constrain image-guided corrections around the pretrained student's predictions to preserve video dynamics and incorporate an optical-flow-based motion reward to improve motion quality and temporal consistency.
Across specialized image experts and multiple video-student backbones, our method consistently outperforms video-teacher OPD baselines, with further studies demonstrating effective transfer across connector designs and heterogeneous architectures.
These results establish image-to-video distillation as an effective route to improving video generation by drawing on the diverse and evolving capabilities of the image-generation ecosystem.

\end{abstract}

\section{Introduction}

\begin{wrapfigure}{r}{0.5\textwidth}
\vspace{-12mm}
    \centering
    \includegraphics[width=0.85\linewidth]{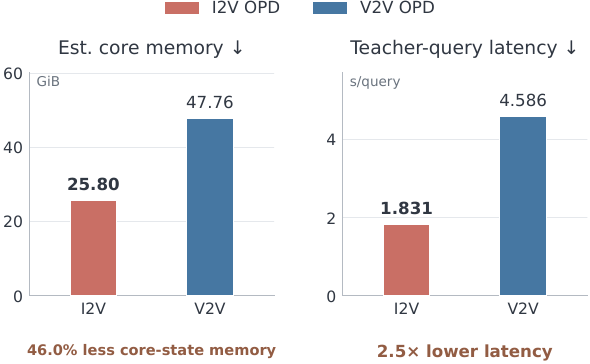}\vspace{-3.5mm}
\caption{
\textbf{Computational costs of I2V and V2V OPD.}
Both train Wan2.1-1.3B using SD3.5-Medium and Wan2.1-14B
teachers, respectively.
Left: estimated student training-state and teacher memory,
excluding activations and auxiliary modules.
Right: measured teacher-query latency.
}\vspace{-4mm}
    \label{fig:teacher_motivation}
\end{wrapfigure}
\begin{figure*}[t]
    \centering
    \includegraphics[width=\textwidth]{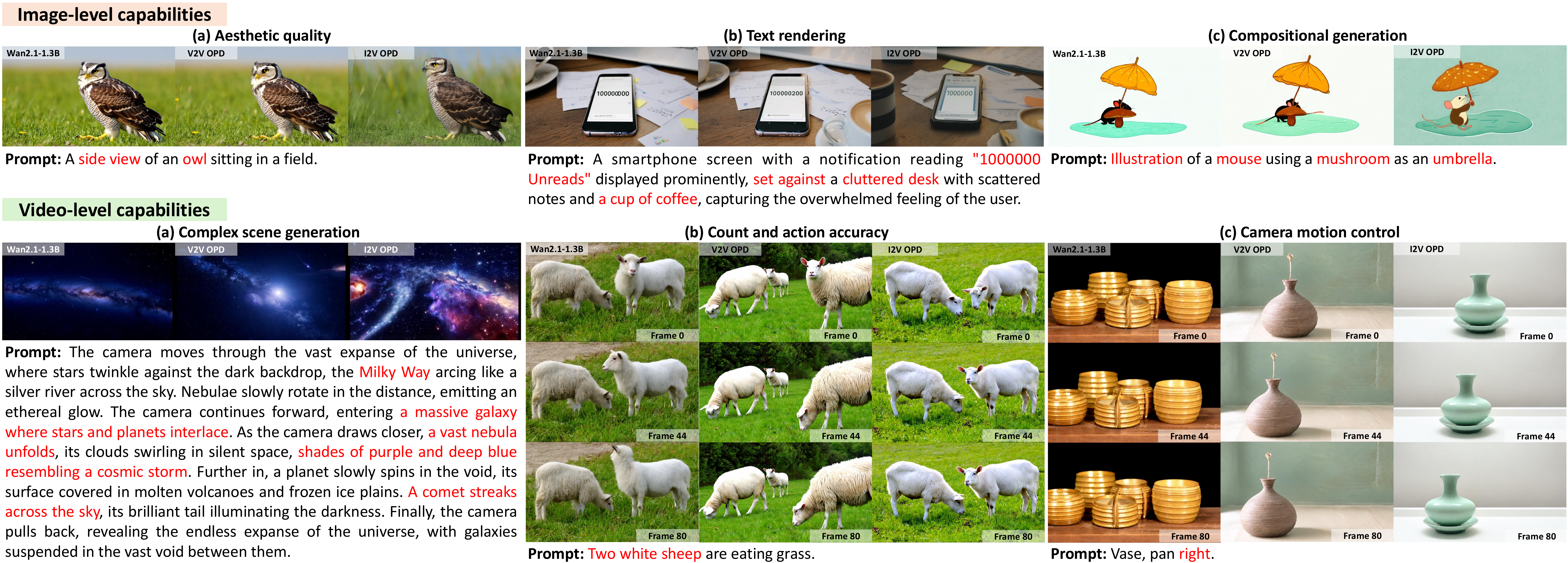}\vspace{-3mm}
\caption{
\textbf{Qualitative comparisons of Base, V2V OPD, and I2V OPD.}
Base denotes the pretrained Wan2.1-1.3B model.
Top: aesthetic quality on DrawBench, text rendering on the
Text Rendering benchmark, and compositional generation.
Bottom: complex scene generation, count and action accuracy,
and camera motion control on VBench-2.0 prompts.
}
    \label{fig:teacher_gap}\vspace{-6mm}
\end{figure*}

Video diffusion models have achieved substantial progress in visual fidelity and temporal coherence
\citep{ho2022video,blattmann2023align,wan2025wan}.
As these models become stronger, an increasingly important practical setting is to adapt a pretrained video model toward specific downstream capabilities, such as improved aesthetics and text rendering, without retraining the model from scratch.
On-policy distillation (OPD) is an effective strategy for this setting which queries the teacher on states sampled from the student's own generation process, thereby steering the student's generation distribution toward that of the stronger teacher during post-training \citep{li2026diffusionopd,fang2026flowopd}.
In the video domain, stronger video diffusion teachers have been shown to improve weaker video models \citep{yin2025causvid,huang2025selfforcing}, for example by accelerating student inference through distillation.
However, OPD training requires repeated evaluations of an expensive video teacher at student-generated states, introducing substantial computational overhead.
As shown in \Cref{fig:teacher_motivation}, video-teacher OPD has an estimated core model-state memory footprint of 47.76 GiB and a mean teacher-query latency of 4.59 seconds under the evaluated configurations.
In contrast, image-teacher OPD reduces these costs to 25.80 GiB and 1.83 seconds, respectively, while image generation models can provide supervision for downstream video tasks such as visual aesthetics and text rendering \citep{guo2024animatediff,kwon2024harivo,chen2024videocrafter2,zhu2024avdm2}.
Qualitative comparisons in \Cref{fig:teacher_gap} further illustrate the potential of image-teacher OPD to improve visual quality and prompt adherence over the pretrained student and video-teacher OPD.

This motivates us to explore whether image generation models post-trained for specialized tasks can serve as attractive teachers for video generation, improving the student video model while maintaining training efficiency. However, applying image models to video generation faces two challenges.
First, image and video models typically use different VAEs, resulting in mismatched latent spaces with different channel dimensions and spatial or temporal compression rates. Thus, the student’s video latents during training cannot be directly processed by the image teacher, nor can the teacher’s predictions serve as targets in the student’s latent space. 
Second, image experts are trained on static images and do not capture temporal dynamics such as cross-frame object and scene evolution, so their frame-level supervision may disrupt temporal coherence. 
These challenges raise a central question:
\textit{How can we transfer specialized image expertise to a video student through on-policy distillation across heterogeneous latent spaces while preserving its learned temporal dynamics?}

To address this question, we propose \textbf{MILD}, a \textbf{M}otion-Preserving \textbf{I}mage-to-Video \textbf{L}atent \textbf{D}istillation framework that transfers specialized visual capabilities from image experts while preserving pretrained video dynamics.
In particular, we first develop an adaptive connector to bridge image and video latent spaces and maintain their alignment as the student evolves throughout training. 
To preserve video dynamics, we anchor image-guided corrections to the student's frozen pretrained predictions and constrain their magnitude and inter-frame variation.
An optical-flow-based motion reward further complements static image expertise with explicit motion feedback, improving motion quality and temporal consistency.
Our main contributions are summarized as follows:
\vspace{-2mm}
\begin{itemize}[leftmargin=*]
    \item We introduce an image-to-video OPD framework, MILD, that transfers specialized visual capabilities from image experts to video students, reducing reliance on costly specialized video teachers.
    Its adaptive connector supports both linear and nonlinear designs and enables transfer across different VAEs and backbone architectures.

    \item We combine image-expert supervision with a frozen video reference and explicit temporal constraints to improve frame quality while preserving pretrained video dynamics.
    An auxiliary motion reward complements image supervision with direct feedback on video motion.

    \item Experiments on Wan2.1-1.3B, Wan2.2-5B, and LTX-Video-2B demonstrate effective capability transfer at lower teacher-query costs on VBench-2.0 and EvalCrafter, while connector ablations and cross-architecture distillation support the framework's generality.
\end{itemize}

\vspace{-2mm}
\section{Related Work}
\label{sec:related_work}
\vspace{-3mm}

\paragraph{Image priors for video generation.}
Prior work extends image models with temporal
modules~\citep{singer2022makeavideo,blattmann2023align,guo2024animatediff},
fine-tunes spatial modules with high-quality
images~\citep{chen2024videocrafter2}, or uses inference-time
denoiser fusion~\citep{shao2025ivmixed} and image-supervised
distillation~\citep{zhai2024motionconsistency}.
We transfer specialized image predictions across heterogeneous
latent spaces, using bounded corrections to a video anchor
to compensate for missing temporal context.

\vspace{-2mm}
\paragraph{On-policy diffusion distillation.}
OPD supervises students along their own generation
trajectories~\citep{lu2025onpolicydistillation,yu2026mismatch}.
DMD2 reduces multi-step training--inference
mismatch~\citep{yin2024improved}; CausVid distills bidirectional
video teachers into autoregressive students, while Self Forcing
reduces exposure bias through self-generated
rollouts~\citep{yin2025causvid,huang2025selfforcing}.
DiffusionOPD and Flow-OPD consolidate specialized image
teachers along student denoising
trajectories~\citep{li2026diffusionopd,fang2026flowopd}.
We transfer image expertise to video by pulling image scores
into video coordinates and integrating video denoising
and motion refinement.

\vspace{-2mm}
\paragraph{Motion-aware video post-training.}
Prior work improves motion through temporal adaptation,
video guidance, or motion-specific
objectives~\citep{zhao2024motiondirector,lee2025videoguide,li2025t2vturbov2,xue2025mogan}.
InstructVideo uses image rewards for video fine-tuning,
while VADER backpropagates reward gradients through video
generation~\citep{yuan2024instructvideo,prabhudesai2407video}.
Our motion feedback corrects image-induced denoising
directions to construct temporally compatible targets
from frame-level supervision.
\vspace{-2mm}
\section{Method}
\label{sec:method}
\vspace{-2mm}

\begin{figure*}[t]
    \centering
    \includegraphics[width=\textwidth]
    {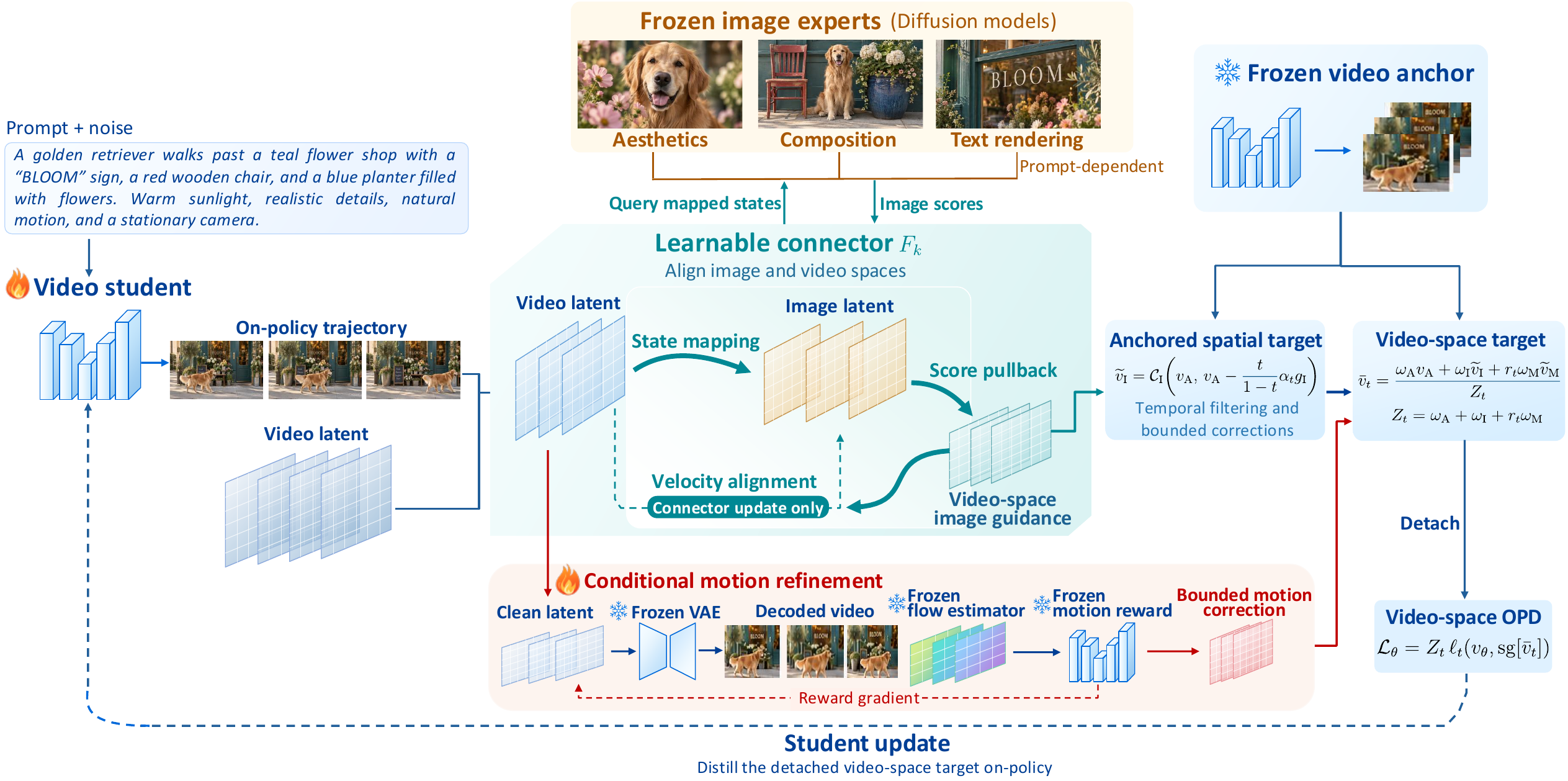}\vspace{-3mm}
     \caption{
    \textbf{Overview of MILD.}
    Learnable connectors transfer image-expert supervision
    to on-policy video states.
    A frozen video anchor bounds spatial corrections,
    while optical-flow feedback provides motion refinement.
    Predictions are fused into a detached distillation target;
    connectors use a separate alignment objective.
    }\vspace{-5mm}
    \label{fig:method_framework}
\end{figure*}

Figure~\ref{fig:method_framework} outlines MILD from frozen image teachers to a video student.
We begin by defining the distillation objective (Section~\ref{sec:method_opd}) and introducing learnable connectors to align heterogeneous representations (Section~\ref{sec:method_connectors}).
Building on this alignment, we constrain and refine image supervision using a video anchor and motion feedback (Section~\ref{sec:method_anchor}).
Finally, we describe target fusion and connector alignment (Section~\ref{sec:method_joint}).

\vspace{-2mm}
\subsection{On-policy distillation objective}
\label{sec:method_opd}
\vspace{-2mm}

We first formulate an on-policy distillation objective that matches the video student's denoising transitions to target transitions at latent states sampled from its current generation trajectories.
Let $v_\theta(x_t,t,c)$ denote the student's velocity at video latent $x_t$, diffusion time $t$, and text condition $c$, and let $\bar v(x_t,t,c)$ denote a target velocity in the same video latent space.
For a reverse step from $t$ to $t-h$, let $\pi_u(\cdot\mid x_t,t,c)$ denote the transition induced by velocity $u$.
The OPD objective is
\begin{equation}
    \mathcal L_{\mathrm{OPD}}
    =
    \mathbb E_{c,t,\,x_t\sim p_\theta(\cdot\mid t,c)}
    \left[
        \ell_t(v_\theta,\bar v)
    \right],
    \quad
    \ell_t(v_\theta,\bar v)
    :=
    \frac{1}{d}
    D_{\mathrm{KL}}\!\left(
        \pi_{v_\theta}(\cdot\mid x_t,t,c)
        \,\Vert\,
        \pi_{\bar v}(\cdot\mid x_t,t,c)
    \right),
    \label{eq:opd_kl}
\end{equation}
where $p_\theta(\cdot\mid t,c)$ is the distribution of states visited by the current student at time $t$, and $d$ is the number of video-latent entries.
Both transitions in the KL divergence are conditioned on the same $(x_t,t,c)$.
We next obtain a closed form for this loss under the reverse-SDE Euler formulation of OPD~\citep{li2026diffusionopd}.
We use the rectified-flow convention $x_t=(1-t)x_0+t\epsilon$, where $\epsilon\sim\mathcal{N}(0,I)$ is independent of $x_0$~\citep{liu2023rectifiedflow}.
For $0<t<1$ and $0<h\leq t$, the transition associated with $u\in\{v_\theta,\bar v\}$ is
$\pi_u=\mathcal{N}(\mu_u,\sigma_t^2hI)$,
where $\mu_u$ is its reverse-SDE Euler mean and $\sigma_t>0$ is the shared diffusion coefficient.
Under this convention, the difference between the transition means is
\[
    \mu_{v_\theta}-\mu_{\bar v}
    =
    -h\left(1+\frac{\sigma_t^2(1-t)}{2t}\right)
    (v_\theta-\bar v).
\]
Since the transitions share the same covariance, their KL divergence depends only on this mean difference. Substitution gives
\begin{align}
    \ell_t(v_\theta,\bar v)
    &=
    \frac{\|\mu_{v_\theta}-\mu_{\bar v}\|_{\mathrm{av}}^2}
         {2\sigma_t^2h}
     =w(t,h)\|v_\theta-\bar v\|_{\mathrm{av}}^2,
    \label{eq:opd_closed_form}\\
    w(t,h)
    &=
    \frac{h}{2\sigma_t^2}
    \left(1+\frac{\sigma_t^2(1-t)}{2t}\right)^2,
    \label{eq:opd_weight}
\end{align}
where $\|q\|_{\mathrm{av}}^2=\|q\|_2^2/d$.
Thus, matching these Gaussian denoising transitions is equivalent to minimizing a weighted velocity error in video coordinates.
The complete reverse-SDE derivation is provided in~\Cref{app:opd}.

\vspace{-2mm}
\subsection{Cross-architecture image guidance transfer}
\label{sec:method_connectors}
\vspace{-2mm}

Video students and image experts use different VAEs and latent layouts, so their latent states and denoising velocities cannot be compared directly. We introduce a differentiable connector $F_\psi:\mathbb R^{d^{\mathrm V}}\to\mathbb R^{d^{\mathrm I}}$ to evaluate student states using image experts, where $d^{\mathrm V}$ and $d^{\mathrm I}$ denote the respective latent dimensions.

\vspace{-2mm}
\paragraph{Transferring image guidance.}
We select video-latent slices at fixed temporal intervals, resize them to the experts' spatial layout, and map them into the image latent space with a learnable transformation. The $K$ frozen experts share a VAE and input layout, allowing a common connector. To compare denoising behavior, we also map the student's velocity into image coordinates. Holding $\psi$ fixed, the chain rule gives
\begin{equation}
    z_t=F_\psi(x_t),
    \qquad
    \widehat v_\theta^{\mathrm I}
    =J_{F_\psi}(x_t)v_\theta(x_t,t,c),
    \label{eq:connector_mapping}
\end{equation}
where $J_{F_\psi}$ is the Jacobian with respect to $x_t$. The mapped velocity $\widehat v_\theta^{\mathrm I}$ can therefore be compared with the expert predictions $u_k(z_t,t,c)$ at the mapped student state.
Expert predictions are expressed in image coordinates and cannot directly serve as video-space updates. Since the connector need not be invertible, we convert the expert velocities into image scores $s_k^{\mathrm I}$ and pull their guidance back to the video latent space through the chain rule:
\begin{equation}
    g_{\mathrm I}
    =
    J_{F_\psi}(x_t)^\top
    \sum_{k=1}^{K}\beta_k(c)s_k^{\mathrm I},
    \qquad
    \beta_k(c)\geq0,\quad \sum_k\beta_k(c)=1.
    \label{eq:score_pullback}
\end{equation}
Here, $g_{\mathrm I}$ gives a video-space direction toward greater agreement with the image experts. The fixed prompt-dependent weights $\beta_k(c)$ select relevant expertise; for example, OCR guidance is disabled when text rendering is not required. 
Appendix~\ref{app:score_pullback} details score conversion and pullback.

\vspace{-2mm}
\paragraph{Aligning the connector.}
Matching latent dimensions alone does not ensure that the mapped video states preserve the visual content or denoising behavior expected by the image experts. We therefore train the connector using content and velocity alignment. For content alignment, we encode the same image with both VAEs: the video VAE receives a static clip of repeated images, while the image VAE's output is replicated across the selected temporal slices. This yields paired clean latents $(\bar x_0,\bar z_0)$. For velocity alignment, we compare the mapped student velocity with expert predictions at states sampled from the student's current trajectories:
\begin{equation}
\begin{aligned}
    \mathcal L_{\mathrm{conn}}
    =
    \lambda_z\mathbb E
    \|F_\psi(\bar x_0)-\bar z_0\|_{\mathrm{av}}^2
    +\frac{\lambda_v}{K}\sum_{k=1}^{K}\mathbb E
    \|\widehat v_\theta^{\mathrm I}
      -u_k(z_t,t,c)\|_{\mathrm{av}}^2
    +\lambda_{\mathrm{reg}}\mathcal R(\psi).
\end{aligned}
\label{eq:connector_objective}
\end{equation}
The content term grounds the mapping in shared visual content, while the velocity term adapts it to the student's evolving denoising trajectories. The regularizer $\mathcal R(\psi)$ penalizes deviations from the initial connector parameters. During connector updates, input latents, student velocities, and expert predictions are detached, so only $\psi$ is optimized. 
Although $z_t$ depends on $F_\psi$, expert predictions are fixed targets; the velocity term updates the connector through $J_{F_\psi}(x_t)v_\theta$.

Our default connector is a bias-free linear map $F_\psi(x)=A_\psi x$, implemented using fixed slice selection and resizing followed by learnable spatial convolutions. Its Jacobian satisfies $J_{F_\psi}(x)v=A_\psi v$, so the same operator maps both states and velocities. Nonlinear connectors use Jacobian--vector products under the same alignment objective. Neither implementation requires an invertible mapping, a shared image--video VAE, or matching backbone architectures. Appendix~\ref{app:connector_geometry} provides the derivation and nonlinear training details.

\vspace{-2mm}
\subsection{Preserving video dynamics and refining motion}
\label{sec:method_anchor}
\vspace{-2mm}

Transferred image guidance can improve spatial content, but it does not account for inter-frame relationships. We therefore constrain its effect using a frozen video reference and optionally add explicit motion feedback.

\vspace{-2mm}
\paragraph{Preserving video dynamics.}
We use the velocity $v_{\mathrm A}$ predicted by a frozen video reference as an anchor and apply the transferred image guidance as a controlled correction:
\begin{equation}
    \widetilde v_{\mathrm I}
    =
    \mathcal C_{\mathrm I}\!\left(
        v_{\mathrm A},
        v_{\mathrm A}
        -\frac{t}{1-t}\alpha_t g_{\mathrm I}
    \right),
    \qquad 0<t<1.
    \label{eq:processed_image_target}
\end{equation}
The factor $t/(1-t)$ converts a score correction into a velocity correction under the rectified-flow convention. The scale $\alpha_t\geq0$ normalizes guidance relative to the anchor score and caps the initial correction. The operator $\mathcal C_{\mathrm I}$ further limits its magnitude relative to the anchor and suppresses abrupt inter-frame variation. 
This produces an image-enhanced target while constraining departures from pretrained video behavior.
Appendix~\ref{app:anchored_target} details the guidance normalization and the temporal correction operator.


\vspace{-2mm}
\paragraph{Refining motion with explicit feedback.}
Constraining image guidance limits temporal disruption but does not directly reward realistic motion. We hence optionally refine the image-enhanced target using a learned motion reward $R$. A frozen RAFT estimator~\citep{teed2020raft} extracts optical flow from the decoded video, and a motion discriminator scores it. The discriminator is trained to distinguish real-video flows from generated and temporally corrupted flows. Specifically, we decode the clean latent prediction $\widehat x_0=x_t-t\operatorname{sg}[\widetilde v_{\mathrm I}]$ and use the reward gradient to refine the target:
\begin{equation}
    \widetilde v_{\mathrm M}
    =
    \mathcal C_{\mathrm M}\!\left(
        \widetilde v_{\mathrm I},
        \widetilde v_{\mathrm I}
        -\gamma_t\nabla_{x_t}
        R\!\left(D^{\mathrm V}(\widehat x_0)\right)
    \right).
    \label{eq:processed_motion_target}
\end{equation}
Here, $D^{\mathrm V}$ is the frozen video decoder, $\operatorname{sg}$ stops gradients through the image target, and $\gamma_t\geq0$ controls the reward correction. Before temporal processing, subtracting this gradient from the velocity moves the clean prediction locally toward higher motion reward. The operator $\mathcal C_{\mathrm M}$ controls the correction magnitude and its inter-frame variation. The decoder, RAFT, and discriminator remain frozen during refinement, while allowing gradients through their inputs. Appendix~\ref{app:implementation_details} describes discriminator training, and Appendix~\ref{app:motion} details reward differentiation and correction scaling.

\vspace{-2mm}
\subsection{Joint optimization}
\label{sec:method_joint}
\vspace{-2mm}

At each student-update state, we combine the frozen video anchor $v_{\mathrm A}$, image-enhanced target $\widetilde v_{\mathrm I}$, and, when applied, motion-refined target $\widetilde v_{\mathrm M}$. Motion refinement requires differentiation through video decoding and optical-flow estimation, so we apply it only at selected states; anchor and image supervision remain active at every state. Let $r_t\in\{0,1\}$ indicate whether motion refinement is applied at $x_t$. For positive weights $\omega_{\mathrm A}$, $\omega_{\mathrm I}$, and $\omega_{\mathrm M}$, define $Z_t=\omega_{\mathrm A}+\omega_{\mathrm I}+r_t\omega_{\mathrm M}$. The fused video-space target is $\bar v_t=Z_t^{-1}(\omega_{\mathrm A}v_{\mathrm A}+\omega_{\mathrm I}\widetilde v_{\mathrm I}+r_t\omega_{\mathrm M}\widetilde v_{\mathrm M})$. Appendix~\ref{app:implementation_details} specifies the diffusion-time range eligible for motion refinement and the per-iteration activation budget.

For $N$ states sampled from the current student trajectories, we optimize the student with
\begin{equation}
    \mathcal L_{\mathrm{student}}
    =\frac{1}{N}\sum_{j=1}^{N}
    Z_{t_j}\,
    \ell_{t_j}\!\left(
        v_\theta(x_{t_j},t_j,c),
        \operatorname{sg}[\bar v_{t_j}]
    \right).
    \label{eq:student_objective}
\end{equation}
Because $\ell_t$ is a weighted squared velocity error, the factor $Z_t$ restores the component weights after target normalization: with the component targets detached, the student gradient equals that from matching each active target separately with its corresponding fusion weight.

Student updates change the on-policy states and velocities used to align the connector. After a connector-only warm-up, each iteration updates the student using \Cref{eq:student_objective} and the connector using \Cref{eq:connector_objective}. The fused target is detached during student updates; input states, student velocities, and image-expert predictions are detached during connector updates. Thus, each objective updates only its intended parameters. The video anchor, image experts, and auxiliary motion networks remain frozen, and inference uses only the updated student. Appendix~\ref{app:optimization} details the two update rules.
\vspace{-2mm}
\section{Experiments}
\vspace{-2mm}

\subsection{Experimental Setting}
\label{sec:experimental_setup}
\vspace{-2mm}

\begin{wraptable}{r}{0.5\columnwidth}
\vspace{-9mm}
\centering
\caption{\textbf{Experimental setup.}
Video-teacher pairings and shared image-expert backbone.
Each V2V configuration uses one video teacher.}
\label{tab:training_configuration}
\footnotesize
\setlength{\tabcolsep}{3pt}
\renewcommand{\arraystretch}{0.95}

\begin{tabular}{@{}lll@{}}
\toprule
\textbf{Video student}
& \textbf{V2V teacher}
& \textbf{Image experts} \\
\midrule
\multirow{3}{*}{Wan2.1-1.3B}
& Wan2.1-14B
& \multirow{5}{*}{SD3.5-Medium} \\
\cmidrule(lr){2-2}
& Wan2.2-5B & \\
\cmidrule(lr){2-2}
& Wan2.2-A14B & \\
\cmidrule(lr){1-2}
Wan2.2-5B & Wan2.2-A14B & \\
\cmidrule(lr){1-2}
LTX-Video-2B & LTX-Video-13B & \\
\bottomrule
\end{tabular}
\vspace{-4mm}
\end{wraptable}

\textbf{Setup.} We evaluate MILD on Wan2.1-1.3B, Wan2.2-5B~\citep{wan2025wan}, and LTX-Video-2B~\citep{hacohen2024ltxvideo}, using a frozen copy of each pretrained student as its video anchor. Following DiffusionOPD~\citep{li2026diffusionopd}, we use three SD3.5-Medium~\citep{esser2024scaling} image experts specializing in compositional generation~\citep{ghosh2023geneval}, aesthetics, and text rendering. Table~\ref{tab:training_configuration} lists the video-teacher pairings and the image-expert backbone shared across students. Ordered as GenEval, aesthetic, and OCR experts, their image-score weights are $(0.5,0.5,0)$ for ordinary prompts and $(0.4,0.4,0.2)$ for text-related prompts. 
We construct a pool of 10,000 training prompts, with 5,000 from OpenVid-1M~\citep{nan2025openvid} and 5,000 from LVD-2M~\citep{xiong2024lvd2m}. Each iteration samples one prompt and generates an on-policy trajectory for an 81-frame video at $480\times832$ resolution. Training comprises 50 connector-alignment iterations followed by 200 joint optimization iterations.
Further implementation details and motion-reward validation are provided in Appendices~\ref{app:implementation_details} and~\ref{app:motion_validation}.

\textbf{Baselines.}
We compare MILD with the pretrained students and two types of V2V OPD baselines: \textbf{V2V OPD (teacher name)} uses a larger pretrained video teacher, while \textbf{V2V OPD (Aes.\ SFT teacher)} uses an aesthetic-SFT video teacher.
To construct the latter, we fine-tune a copy of each student backbone on video--text data from
AesVideo-Bench~\citep{han2026aesrm}, then freeze the resulting model as its V2V OPD teacher.
Table~\ref{tab:training_configuration} lists the video-teacher pairings, including Wan2.2-5B and Wan2.2-A14B teachers for Wan2.1-1.3B to assess transfer across video model variants.
We evaluate video quality using VBench-2.0~\citep{zheng2025vbench2} and EvalCrafter~\citep{liu2024evalcrafter}.

\begin{table*}[t]
\centering
\caption{\textbf{Evaluation results on VBench-2.0 and EvalCrafter.} MILD outperforms both V2V OPD baselines on the aggregate scores of both benchmarks across all three student backbones. Compared with the larger-teacher V2V OPD baseline, MILD reduces the sum of per-GPU peak allocated memory by 39.32\%, 64.07\%, and 19.53\% on Wan2.1-1.3B, Wan2.2-5B, and LTX-Video-2B, respectively (Figure~\ref{fig:gpu_memory_comparison}). Bold and underlined values indicate the best and second-best results within each student group.}
\label{tab:video_benchmarks}
\footnotesize
\setlength{\tabcolsep}{3pt}
\renewcommand{\arraystretch}{1}
\resizebox{\textwidth}{!}{%
\begin{tabular}{@{}lccccc c@{\hspace{9pt}}cccc c@{}}
\toprule
& \multicolumn{6}{c}{\textbf{VBench-2.0 $\uparrow$}}
& \multicolumn{5}{c}{\textbf{EvalCrafter $\uparrow$}} \\
\cmidrule(lr){2-7}
\cmidrule(l){8-12}
\textbf{Method}
& \textbf{Creativity}
& \textbf{Commonsense}
& \textbf{Controllability}
& \shortstack{\textbf{Human}\\\textbf{Fidelity}}
& \textbf{Physics}
& \textbf{AVG.}
& \shortstack{\textbf{Visual}\\\textbf{Quality}}
& \shortstack{\textbf{Text--Video}\\\textbf{Alignment}}
& \shortstack{\textbf{Motion}\\\textbf{Quality}}
& \shortstack{\textbf{Temporal}\\\textbf{Consistency}}
& \shortstack{\textbf{Final Sum}\\\textbf{Score}} \\
\midrule
\multicolumn{12}{@{}l}{\textit{Large video teachers (reference)}} \\
Wan2.1-14B
& 48.91 & 59.95 & 27.44 & 90.94 & 44.09 & 54.27
& 66.25 & 56.78 & 53.97 & 63.93 & 240.93 \\
Wan2.2-A14B
& 52.98 & 67.73 & 35.36 & 82.59 & 51.60 & 58.05
& 66.47 & 60.22 & 54.58 & 63.10 & 244.37 \\
LTX-Video-13B
& 55.02 & 40.84 & 13.90 & 87.60 & 43.14 & 48.10
& 64.08 & 51.99 & 56.00 & 65.04 & 237.11 \\
\midrule
\multicolumn{12}{@{}l}{\textit{Wan2.1-1.3B}} \\
Base
& 45.86 & 59.68 & 22.85 & 85.16 & 42.07 & 51.12
& 65.78 & 56.51 & 53.36 & \underline{63.22} & 238.87 \\
V2V OPD (Wan2.1-14B)
& 47.34 & \underline{63.71} & 24.67 & 85.17 & \underline{45.32} & \underline{53.24}
& \underline{66.64} & 56.38 & 53.70 & 63.03 & 239.75 \\
V2V OPD (Aes.\ SFT teacher)
& \textbf{48.12} & 61.13 & \underline{24.94} & \textbf{86.45} & 44.94 & 53.12
& 65.47 & \textbf{58.62} & \underline{54.45} & 63.16 & \underline{241.70} \\
\rowcolor{gray!10}
\textbf{MILD}
& \underline{47.89} & \textbf{64.56} & \textbf{25.17} & \underline{85.47} & \textbf{46.52} & \textbf{53.92}
& \textbf{66.84} & \underline{57.89} & \textbf{54.58} & \textbf{63.79} & \textbf{243.10} \\
\midrule
\multicolumn{12}{@{}l}{\textit{Wan2.2-5B}} \\
Base
& 48.20 & 58.50 & 20.02 & 81.89 & 49.31 & 51.58
& 61.48 & 55.63 & 54.00 & 61.93 & 233.04 \\
V2V OPD (Wan2.2-A14B)
& 48.29 & 60.81 & \underline{20.79} & 81.98 & \underline{51.60} & 52.69
& 62.06 & 57.51 & 54.07 & \textbf{63.10} & 236.74 \\
V2V OPD (Aes.\ SFT teacher)
& \underline{48.64} & \underline{61.69} & 19.07 & \underline{83.75} & 50.62 & \underline{52.75}
& \underline{62.23} & \underline{57.79} & \underline{54.11} & 62.73 & \underline{236.86} \\
\rowcolor{gray!10}
\textbf{MILD}
& \textbf{49.50} & \textbf{62.84} & \textbf{21.80} & \textbf{84.89} & \textbf{54.29} & \textbf{54.66}
& \textbf{63.76} & \textbf{59.12} & \textbf{54.43} & \underline{62.79} & \textbf{240.10} \\
\midrule
\multicolumn{12}{@{}l}{\textit{LTX-Video-2B}} \\
Base
& 47.95 & 43.72 & 12.17 & 67.35 & 37.77 & 41.79
& 56.31 & 46.78 & 54.96 & 62.76 & 220.81 \\
V2V OPD (LTX-13B)
& 56.68 & 33.38 & \underline{16.22} & 68.82 & 45.43 & 44.11
& 56.50 & 47.05 & 55.19 & \underline{64.39} & 223.13 \\
V2V OPD (Aes.\ SFT teacher)
& \underline{56.75} & \underline{45.36} & 15.97 & \underline{69.15} & \underline{47.14} & \underline{46.87}
& \textbf{56.62} & \underline{47.61} & \underline{55.21} & 64.21 & \underline{223.65} \\
\rowcolor{gray!10}
\textbf{MILD}
& \textbf{57.23} & \textbf{48.85} & \textbf{16.78} & \textbf{72.38} & \textbf{50.06} & \textbf{49.06}
& \underline{56.58} & \textbf{48.41} & \textbf{55.30} & \textbf{64.51} & \textbf{224.80} \\
\bottomrule
\end{tabular}
}\vspace{-6mm}
\end{table*}


\begin{figure}[t]
\centering
\includegraphics[width=1\columnwidth]{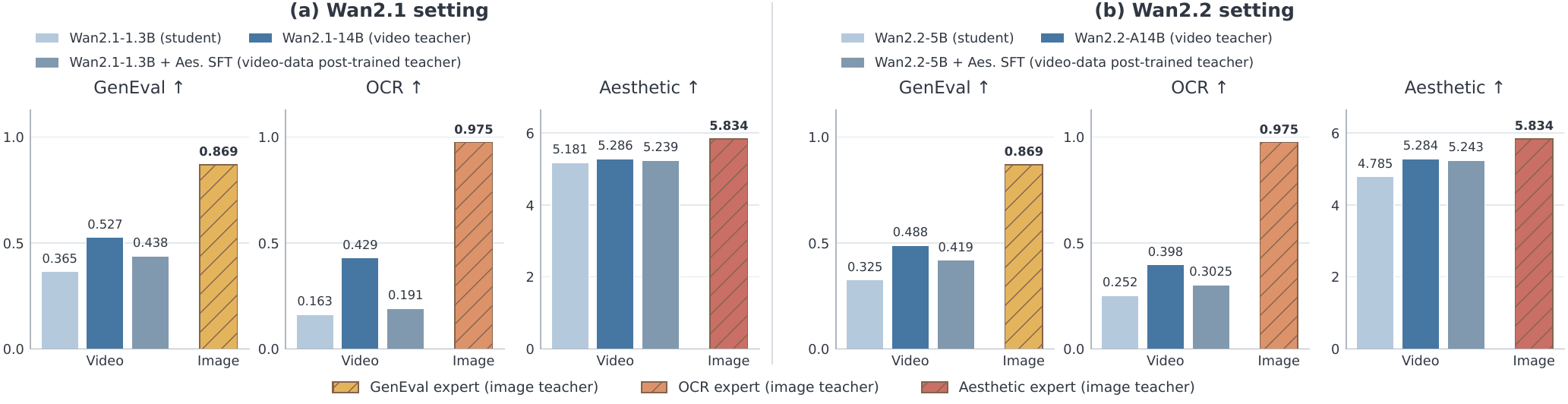}
\vspace{-5mm}
\caption{
\textbf{Spatial capability comparisons across Wan2.1 and Wan2.2 settings.}
Each setting compares the student, a larger video teacher,
a video teacher post-trained on aesthetic video data,
and specialized image teachers.
Image experts achieve the highest GenEval, OCR, and
Aesthetic scores in both settings.
}
\label{fig:teacher_capabilities}
\vspace{-6mm}
\end{figure}


\begin{table*}[!t]
\centering
\caption{
\textbf{Cross-architecture video-to-video OPD on VBench-2.0.}
All OPD variants use Wan2.1-1.3B as the student, with the teacher specified in parentheses.
Distillation from Wan2.2 teachers improves the student's overall performance, showing that the proposed connector transfers supervision between heterogeneous video diffusion models.
}
\label{tab:cross_model_v2v}
\footnotesize
\setlength{\tabcolsep}{5pt}
\renewcommand{\arraystretch}{1}

\resizebox{0.90\textwidth}{!}{%
\begin{tabular}{@{}lcccccc@{}}
\toprule
\textbf{Method}
& \textbf{Creativity}
& \textbf{Commonsense}
& \textbf{Controllability}
& \textbf{Human Fidelity}
& \textbf{Physics}
& \textbf{AVG.} \\
\midrule

Wan2.1-1.3B
& 45.86
& 59.68
& 22.85
& 85.16
& 42.07
& 51.12 \\

V2V OPD (Wan2.1-14B)
& 47.34
& 63.71
& 24.67
& 85.17
& 45.32
& 53.24 \\

V2V OPD (Wan2.2-A14B)
& 48.74
& 65.72
& 24.59
& 82.37
& 42.91
& 52.87 \\

V2V OPD (Wan2.2-5B)
& 48.29
& 58.50
& 23.70
& 82.51
& 43.14
& 51.23 \\

\bottomrule
\end{tabular}\vspace{-3mm}
}
\end{table*}

\vspace{-2mm}
\subsection{Main Results}
\label{sec:main_results}
\vspace{-2mm}

\begin{wrapfigure}{r}{0.5\textwidth}
\vspace{-5mm}
    \centering
    \includegraphics[width=\linewidth]{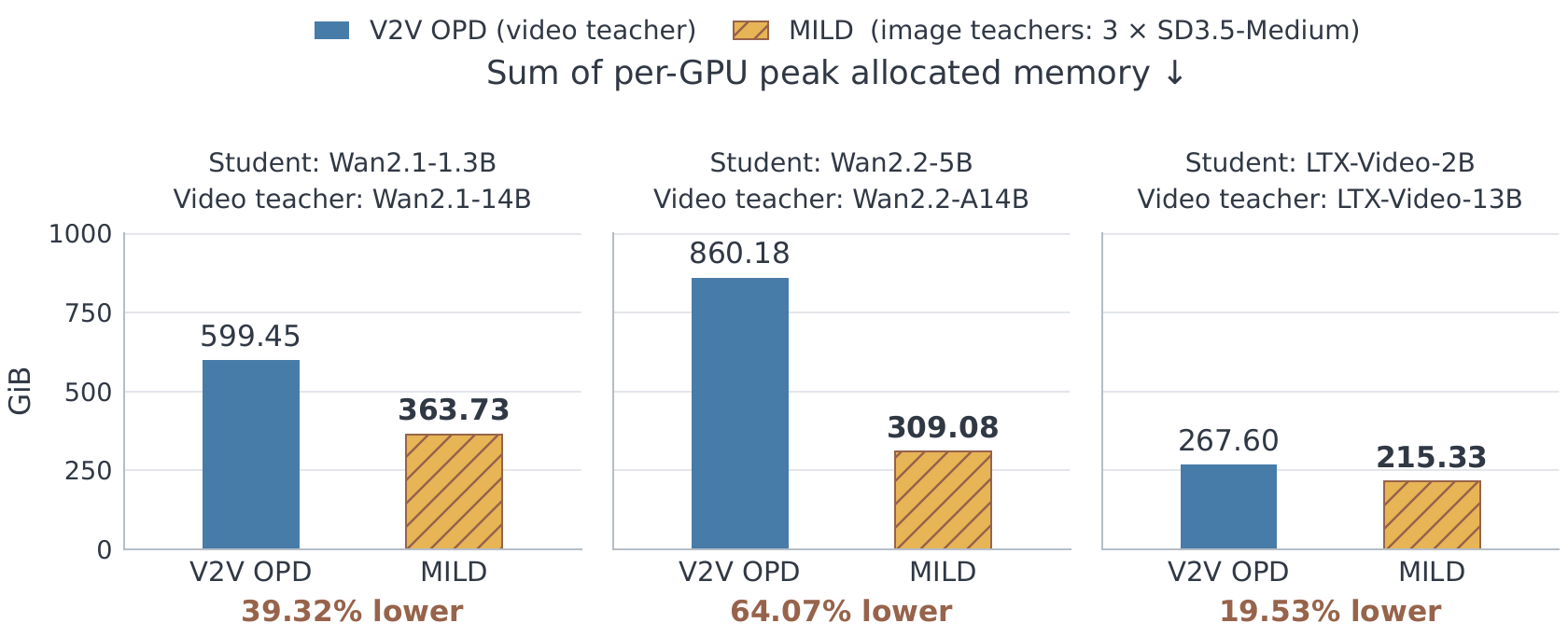}\vspace{-4mm}
    \caption{\textbf{GPU memory comparison across student backbones.} MILD reduces the sum of per-GPU peak allocated memory relative to V2V OPD across all three backbones. The metric sums each participating GPU's peak allocated memory.}\vspace{-4mm}
    \label{fig:gpu_memory_comparison}
\end{wrapfigure}

\textbf{Quantitative Analysis.} 
Table~\ref{tab:video_benchmarks} compares MILD with the pretrained students and V2V OPD baselines across three student backbones on VBench-2.0 and EvalCrafter. MILD achieves stronger aggregate benchmark performance while reducing GPU memory usage, as shown in Figure~\ref{fig:gpu_memory_comparison}. On Wan2.1-1.3B, Wan2.2-5B, and LTX-Video-2B, MILD achieves VBench-2.0 averages of 53.92, 54.66, and 49.06, exceeding the best V2V OPD results of 53.24, 52.75, and 46.87, respectively. It also surpasses both V2V OPD baselines in EvalCrafter final score across all three backbones. Meanwhile, the sum of per-GPU peak allocated memory is reduced by 39.32\%, 64.07\%, and 19.53\%, respectively, relative to the V2V OPD baseline. These results demonstrate that specialized image-expert supervision can outperform video-to-video distillation while requiring less aggregate GPU memory. 
Figure~\ref{fig:teacher_capabilities} compares specialized image experts with students, larger video teachers, and video teachers post-trained on aesthetic video data across the Wan2.1 and Wan2.2 settings.
The GenEval and OCR experts score 0.869 and 0.975, respectively, compared with 0.527 and 0.429 for Wan2.1-14B and 0.488 and 0.398 for Wan2.2-A14B.
The aesthetic expert scores 5.834, exceeding all evaluated video models, whose highest score is 5.286.
These capability advantages support using specialized image experts as sources of spatial supervision for video post-training.
Beyond appearance, MILD improves motion quality, temporal consistency, controllability, and physics over each pretrained student, indicating that transferring static image expertise
is compatible with improved temporal behavior. 
Table~\ref{tab:cross_model_v2v} further demonstrates the connector's applicability to cross-architecture video-to-video OPD. 
Distilling Wan2.2-A14B or Wan2.2-5B into Wan2.1-1.3B raises its VBench-2.0 average from 51.12 to 52.87 or 51.23, respectively. 
This extension shows that the connector provides a supervision interface for heterogeneous diffusion models beyond the image-to-video setting. 
Architectural differences, particularly between the mixture-of-experts Wan2.2-A14B teacher and the dense student, may affect transfer effectiveness; understanding and mitigating these effects remains future work.
Appendix~\ref{app:aes_teacher_comparison} provides further comparisons with video models fine-tuned on aesthetic video data, showing that MILD retains an overall advantage and supporting the effectiveness of transferring specialized image expertise to video generation.

\begin{figure*}[t]
\centering
\includegraphics[width=\textwidth]{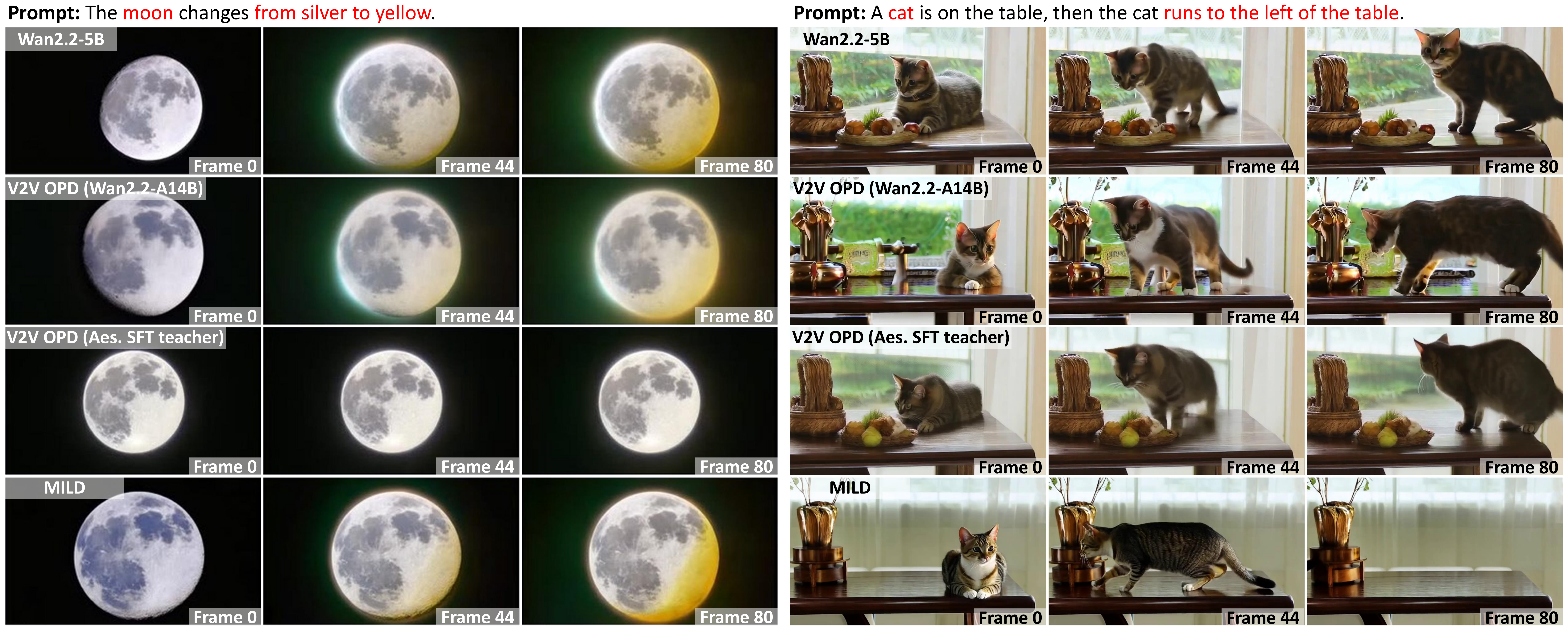}\vspace{-3mm}
\caption{
\textbf{Qualitative comparisons using Wan2.2-5B as the student.} Frames 0, 44, and 80 illustrate MILD's ability to accurately follow prompts involving color changes and directional motion. In contrast, baselines fail to respond correctly to the moon’s color change and produces artifacts (left), while the cat’s motion direction is incorrect (right). More visualizations are in Appendix~\ref{app:additional_cases}.
}
\vspace{-5mm}
\label{fig:qualitative_cases}
\end{figure*}

\vspace{-1mm}
\textbf{Qualitative Analysis.} 
Figure~\ref{fig:qualitative_cases} compares frames 0, 44, and 80 from videos generated with Wan2.2-5B as the student.
For the moon prompt, MILD shows a clear transition toward yellow while retaining visible lunar texture, whereas V2V OPD with the aesthetic SFT teacher shows little color change.
For the cat prompt, MILD depicts the cat moving left across the table and leaving the frame by the final sampled frame, while the compared baselines keep the cat visible on the table.
These cases suggest that MILD transfers spatial expertise while retaining the ability to generate prompt-aligned temporal changes.
Appendix~\ref{app:additional_cases} provides additional cases using Wan2.2-5B as the student, while Appendix~\ref{app:aesthetic_teacher_comparison} compares the aesthetic image expert with the video teacher post-trained on aesthetic video data to illustrate differences in visual
expertise available for distillation.

\vspace{-3.5mm}
\subsection{Spatial Capability Transfer}
\label{sec:spatial_transfer}
\vspace{-3mm}

\begin{table*}[t]
\centering
\caption{\textbf{Frame-level spatial capability evaluation using Wan2.2-5B.} We evaluate the middle frame of each generated video, using DrawBench prompts for the five model-based metrics and FlowGRPO evaluation splits for GenEval and OCR, consistent with DiffusionOPD~\citep{li2026diffusionopd}. MILD achieves the highest score on every metric among the compared student variants.}
\label{tab:drawbench_wan22}
\footnotesize
\setlength{\tabcolsep}{5pt}
\renewcommand{\arraystretch}{1}
\setlength{\aboverulesep}{1pt}
\setlength{\belowrulesep}{1pt}
\resizebox{\textwidth}{!}{%
\begin{tabular}{@{}lccccc@{\hspace{9pt}}cc@{}}
\toprule
& \multicolumn{5}{c}{\textbf{DrawBench}}
& \multicolumn{2}{c}{\textbf{FlowGRPO splits}} \\
\cmidrule(lr){2-6}
\cmidrule(l){7-8}
\textbf{Method}
& \textbf{Aesthetic}
& \textbf{PickScore}
& \textbf{HPSv2}
& \textbf{CLIPScore}
& \textbf{ImageReward}
& \textbf{GenEval}
& \textbf{OCR} \\
\midrule
Base
& 4.7855 & 0.7841 & 0.2293 & 0.2372 & $-0.4390$ & 0.3252 & 0.2522 \\
V2V OPD (Aes.\ SFT teacher)
& \underline{4.9169} & \underline{0.7931} & \underline{0.2321} & \underline{0.2414} & $\underline{-0.3098}$ & 0.4023 & \underline{0.2617} \\
V2V OPD (Wan2.2-A14B)
& 4.8217 & 0.7819 & 0.2253 & 0.2338 & $-0.4610$ & \underline{0.4067} & 0.2556 \\
\rowcolor{gray!10}
\textbf{MILD}
& \textbf{5.0305} & \textbf{0.8111} & \textbf{0.2470} & \textbf{0.2438} & $\mathbf{0.0320}$ & \textbf{0.4454} & \textbf{0.2697} \\
\bottomrule
\end{tabular}
}\vspace{-5.5mm}
\end{table*}

We evaluate spatial capability transfer using the evaluation splits and metrics adopted by DiffusionOPD~\citep{li2026diffusionopd}.
For each evaluation prompt, we first generate a video and extract its middle frame for image-level assessment, using the same frame-selection protocol across all methods.
On DrawBench~\citep{saharia2022photorealistic}, we evaluate Aesthetic~\citep{schuhmann2022laion}, PickScore~\citep{kirstain2023pick}, HPSv2~\citep{wu2023human}, CLIPScore~\citep{hessel2021clipscore}, and ImageReward~\citep{xu2023imagereward} to assess visual aesthetics, predicted human preference, and text--image alignment.
For compositional generation and text rendering, we use the GenEval~\citep{ghosh2023geneval} and OCR evaluation splits provided by FlowGRPO~\citep{liu2026flow}, applying their respective scoring procedures to the extracted middle frames.
As shown in Table~\ref{tab:drawbench_wan22}, MILD leads the compared student variants on all seven frame-level metrics. It achieves a PickScore of 0.8111, an ImageReward score of 0.0320, and an OCR score of 0.2697; its GenEval score also exceeds the strongest V2V OPD result (0.4454 versus 0.4067). These results support the transfer of specialized image expertise to video generation and complement the video-level evaluations of motion quality and temporal consistency. Qualitative comparisons are provided in Appendix~\ref{app:spatial_qualitative}.




\begin{table*}[t]
\centering
\caption{
\textbf{Ablation study on Wan2.2-5B using VBench-2.0.}
The full model uses auxiliary motion refinement, joint
optimization, the default connector, and prompt-dependent
image-score weights that enable OCR guidance only for
text-related prompts.
Single-expert variants use only the indicated image expert.
}
\label{tab:wan22_ablation}
\small
\setlength{\tabcolsep}{5pt}
\renewcommand{\arraystretch}{1.10}

\resizebox{0.96\textwidth}{!}{%
\begin{tabular}{@{}lrrrrrr@{}}
\toprule
\textbf{Variant}
& \textbf{Creativity}
& \textbf{Commonsense}
& \textbf{Controllability}
& \textbf{Human Fidelity}
& \textbf{Physics}
& \textbf{AVG.} \\
\midrule

\multicolumn{7}{@{}l}{
\textit{Pretrained model (reference)}
} \\

Wan2.2-5B (Base)
& 48.20
& 58.50
& 20.02
& 81.89
& 49.31
& 51.58 \\

\midrule

\multicolumn{7}{@{}l}{
\textit{Supervision and optimization strategy}
} \\

w/o image experts
& 45.97
& 60.79
& \underline{20.48}
& 82.11
& 50.95
& 52.06 \\

w/o motion refinement
& \underline{48.64}
& \underline{61.75}
& 19.73
& 83.45
& 53.68
& 53.45 \\

V2V OPD (Wan2.2-A14B) + motion refinement
& 46.86
& 61.68
& 20.40
& 83.61
& 52.74
& 53.06 \\

Alternating optimization
& 48.42
& \textbf{62.84}
& 20.34
& \underline{83.83}
& \underline{53.93}
& \underline{53.87} \\

\rowcolor{gray!10}
\textbf{MILD (joint optimization)}
& \textbf{49.50}
& \textbf{62.84}
& \textbf{21.80}
& \textbf{84.89}
& \textbf{54.29}
& \textbf{54.66} \\

\midrule

\multicolumn{7}{@{}l}{
\textit{Connector architecture}
} \\

$1{\times}1$ linear connector
& 48.20
& 60.80
& 19.62
& 83.75
& 52.37
& 52.95 \\

Nonlinear connector
& \underline{48.74}
& \underline{61.39}
& \underline{19.68}
& \underline{84.33}
& \underline{52.78}
& \underline{53.38} \\

\rowcolor{gray!10}
\textbf{MILD (default connector)}
& \textbf{49.50}
& \textbf{62.84}
& \textbf{21.80}
& \textbf{84.89}
& \textbf{54.29}
& \textbf{54.66} \\

\midrule

\multicolumn{7}{@{}l}{
\textit{Expert selection, weighting, and routing}
} \\

GenEval expert only
& 47.99
& \underline{61.38}
& 19.53
& 83.12
& 49.78
& 52.36 \\

Aesthetic expert only
& \underline{49.26}
& 61.09
& 19.70
& \underline{84.52}
& 51.58
& \underline{53.23} \\

Uniform weights $(1/3,\,1/3,\,1/3)$
& 48.68
& 59.66
& \underline{20.01}
& 83.16
& \underline{53.21}
& 52.94 \\

\rowcolor{gray!10}
\textbf{MILD (prompt-dependent weights)}
& \textbf{49.50}
& \textbf{62.84}
& \textbf{21.80}
& \textbf{84.89}
& \textbf{54.29}
& \textbf{54.66} \\

\bottomrule
\end{tabular}
}\vspace{-6mm}
\end{table*}

\vspace{-3mm}
\subsection{Ablation Study}
\label{sec:ablation_study}
\vspace{-3mm}

We ablate image guidance, motion refinement, optimization strategy, connector architecture, and expert configuration on Wan2.2-5B using VBench-2.0, keeping all other settings fixed unless otherwise stated. Results are reported in Table~\ref{tab:wan22_ablation}. 

\vspace{-1mm}
\noindent\textbf{Supervision and optimization strategy.} We remove image-expert guidance while retaining motion feedback, or remove motion refinement while retaining image guidance; both variants retain the video anchor. The full model achieves an average score of 54.66, compared with 52.06 without image experts and 53.45 without motion refinement, and outperforms both variants across all five categories. It also exceeds V2V OPD with Wan2.2-A14B under the same motion-refinement procedure (53.06), supporting the effectiveness of image-expert transfer when both approaches receive motion feedback. Joint optimization outperforms an alternating schedule of one connector update followed by 16 student updates (54.66 versus 53.87), consistent with the benefit of adapting the connector as the student policy evolves. Both schedules retain the separate objectives and gradient-detachment rules in Section~\ref{sec:method_joint}.

\vspace{-1mm}
\noindent\textbf{Connector architecture.} We replace the default spatial linear connector with a $1{\times}1$ linear or nonlinear connector under the same alignment objective. Linear connectors apply the same operator to states and velocities, while the nonlinear variant uses Jacobian--vector products. The $1{\times}1$ linear and nonlinear variants score 52.95 and 53.38, respectively, both above the pretrained student's 51.58 but below the default connector's 54.66. These results show that the framework supports different connector architectures, while the default design performs best among those tested.

\vspace{-1mm}
\noindent\textbf{Expert selection, weighting, and activation.} We compare the default multi-expert policy with GenEval-only, aesthetic-only, and uniformly weighted variants, keeping the image-guidance scales and target-fusion weights fixed. With experts ordered as GenEval, aesthetic, and OCR, these variants use weights $(1,0,0)$, $(0,1,0)$, and $(1/3,1/3,1/3)$, respectively, for every prompt. The default policy scores 54.66, compared with 53.23 for the stronger single-expert variant and 52.94 for uniform weighting. This comparison supports the combined use of specialized experts with prompt-dependent weighting and activation.

\vspace{-4mm}
\section{Conclusion}
\label{sec:conclusion}
\vspace{-3mm}

We presented MILD for transferring specialized image expertise to video diffusion models through on-policy distillation.
Learnable connectors align heterogeneous latent representations, while video-anchor constraints and motion feedback support spatial improvements alongside temporal consistency.
Experiments across three video backbones demonstrate effective capability transfer and consistent aggregate gains over video-teacher OPD baselines.
Our results establish that image experts can be a practical source of diverse supervision for video post-training.

\newpage

\bibliography{iclr2027_conference}

@article{ho2022video,
  title={Video diffusion models},
  author={Ho, Jonathan and Salimans, Tim and Gritsenko, Alexey and Chan, William and Norouzi, Mohammad and Fleet, David J},
  journal={Advances in neural information processing systems},
  volume={35},
  pages={8633--8646},
  year={2022}
}

@inproceedings{blattmann2023align,
  title={Align your latents: High-resolution video synthesis with latent diffusion models},
  author={Blattmann, Andreas and Rombach, Robin and Ling, Huan and Dockhorn, Tim and Kim, Seung Wook and Fidler, Sanja and Kreis, Karsten},
  booktitle={IEEE/CVF Conference on Computer Vision and Pattern Recognition (CVPR)},
  pages={22563--22575},
  year={2023},
  organization={IEEE}
}

@article{wan2025wan,
  title={Wan: Open and advanced large-scale video generative models},
  author={Wan, Team and Wang, Ang and Ai, Baole and Wen, Bin and Mao, Chaojie and Xie, Chen-Wei and Chen, Di and Yu, Feiwu and Zhao, Haiming and Yang, Jianxiao and others},
  journal={arXiv preprint arXiv:2503.20314},
  year={2025}
}

@article{li2026diffusionopd,
  title={DiffusionOPD: A unified perspective of on-policy distillation in diffusion models},
  author={Li, Quanhao and Yu, Junqiu and Jiang, Kaixun and Wei, Yujie and Xing, Zhen and Li, Pandeng and Chu, Ruihang and Zhang, Shiwei and Liu, Yu and Wu, Zuxuan},
  journal={arXiv preprint arXiv:2605.15055},
  year={2026}
}

@article{fang2026flowopd,
  title={Flow-opd: On-policy distillation for flow matching models},
  author={Fang, Zhen and Huang, Wenxuan and Zeng, Yu and Zhao, Yiming and Chen, Shuang and Feng, Kaituo and Lin, Yunlong and Chen, Lin and Chen, Zehui and Cao, Shaosheng and others},
  journal={arXiv preprint arXiv:2605.08063},
  year={2026}
}

@inproceedings{yin2025causvid,
  title={From slow bidirectional to fast autoregressive video diffusion models},
  author={Yin, Tianwei and Zhang, Qiang and Zhang, Richard and Freeman, William T and Durand, Fredo and Shechtman, Eli and Huang, Xun},
  booktitle={2025 IEEE/CVF Conference on Computer Vision and Pattern Recognition (CVPR)},
  pages={22963--22974},
  year={2025},
  organization={IEEE}
}

@article{huang2025selfforcing,
  title={Self forcing: Bridging the train-test gap in autoregressive video diffusion},
  author={Huang, Xun and Li, Zhengqi and He, Guande and Zhou, Mingyuan and Shechtman, Eli},
  journal={Advances in Neural Information Processing Systems},
  volume={38},
  pages={167283--167308},
  year={2026}
}

@inproceedings{esser2024scaling,
  title={Scaling rectified flow transformers for high-resolution image synthesis},
  author={Esser, Patrick and Kulal, Sumith and Blattmann, Andreas and Entezari, Rahim and M{\"u}ller, Jonas and Saini, Harry and Levi, Yam and Lorenz, Dominik and Sauer, Axel and Boesel, Frederic and others},
  booktitle={International conference on machine learning},
  year={2024}
}

@article{guo2024animatediff,
  title={Animatediff: Animate your personalized text-to-image diffusion models without specific tuning},
  author={Guo, Yuwei and Yang, Ceyuan and Rao, Anyi and Liang, Zhengyang and Wang, Yaohui and Qiao, Yu and Agrawala, Maneesh and Lin, Dahua and Dai, Bo},
  journal={arXiv preprint arXiv:2307.04725},
  year={2023}
}

@inproceedings{kwon2024harivo,
  title={Harivo: Harnessing text-to-image models for video generation},
  author={Kwon, Mingi and Oh, Seoung Wug and Zhou, Yang and Liu, Difan and Lee, Joon-Young and Cai, Haoran and Liu, Baqiao and Liu, Feng and Uh, Youngjung},
  booktitle={European Conference on Computer Vision},
  pages={19--36},
  year={2024},
  organization={Springer}
}

@inproceedings{chen2024videocrafter2,
  title={Videocrafter2: Overcoming data limitations for high-quality video diffusion models},
  author={Chen, Haoxin and Zhang, Yong and Cun, Xiaodong and Xia, Menghan and Wang, Xintao and Weng, Chao and Shan, Ying},
  booktitle={IEEE/CVF Conference on Computer Vision and Pattern Recognition (CVPR)},
  pages={7310--7320},
  year={2024},
  organization={IEEE}
}

@article{zhu2024avdm2,
  title={Accelerating video diffusion models via distribution matching},
  author={Zhu, Yuanzhi and Yan, Hanshu and Yang, Huan and Zhang, Kai and Li, Junnan},
  journal={arXiv preprint arXiv:2412.05899},
  year={2024}
}

@article{zhai2024motionconsistency,
  title={Motion consistency model: Accelerating video diffusion with disentangled motion-appearance distillation},
  author={Zhai, Yuanhao and Lin, Kevin and Yang, Zhengyuan and Li, Linjie and Wang, Jianfeng and Lin, Chung-Ching and Doermann, David and Yuan, Junsong and Wang, Lijuan},
  journal={Advances in Neural Information Processing Systems},
  volume={37},
  pages={111000--111021},
  year={2024}
}

@inproceedings{shao2025ivmixed,
  title={Iv-mixed sampler: Leveraging image diffusion models for enhanced video synthesis},
  author={Shao, Shitong and Bai, Lichen and Xiong, Haoyi and Xie, Zeke and others},
  booktitle={International Conference on Learning Representations},
  volume={2025},
  pages={12396--12424},
  year={2025}
}

@inproceedings{li2025t2vturbov2,
  title={T2v-turbo-v2: Enhancing video model post-training through data, reward, and conditional guidance design},
  author={Li, Jiachen and Long, Qian and Zheng, Jian Skyler and Gao, Xiaofeng and Piramuthu, Robinson and Chen, Wenhu and Wang, William},
  booktitle={International Conference on Learning Representations},
  volume={2025},
  pages={92279--92305},
  year={2025}
}

@inproceedings{zhao2024motiondirector,
  title={Motiondirector: Motion customization of text-to-video diffusion models},
  author={Zhao, Rui and Gu, Yuchao and Wu, Jay Zhangjie and Zhang, David Junhao and Liu, Jia-Wei and Wu, Weijia and Keppo, Jussi and Shou, Mike Zheng},
  booktitle={European Conference on Computer Vision},
  pages={273--290},
  year={2024},
  organization={Springer}
}

@inproceedings{lee2025videoguide,
  title={Videoguide: Improving video diffusion models without training through a teacher’s guide},
  author={Lee, Dohun and Kim, Bryan Sangwoo and Park, Geon Yeong and Ye, Jong Chul},
  booktitle={IEEE/CVF Conference on Computer Vision and Pattern Recognition (CVPR)},
  pages={2599--2608},
  year={2025},
  organization={IEEE}
}

@article{xue2025mogan,
  title={MoGAN: Improving Motion Quality in Video Diffusion via Few-Step Motion Adversarial Post-Training},
  author={Xue, Haotian and Chen, Qi and Wang, Zhonghao and Huang, Xun and Shechtman, Eli and Xie, Jinrong and Chen, Yongxin},
  journal={arXiv preprint arXiv:2511.21592},
  year={2025}
}

@inproceedings{liu2023rectifiedflow,
title={Flow Straight and Fast: Learning to Generate and Transfer Data with Rectified Flow},
author={Xingchao Liu and Chengyue Gong and qiang liu},
booktitle={International Conference on Learning Representations},
year={2023},
}

@inproceedings{teed2020raft,
  title={Raft: Recurrent all-pairs field transforms for optical flow},
  author={Teed, Zachary and Deng, Jia},
  booktitle={European conference on computer vision},
  pages={402--419},
  year={2020},
  organization={Springer}
}

@inproceedings{mescheder2018training,
  title={Which training methods for GANs do actually converge?},
  author={Mescheder, Lars and Geiger, Andreas and Nowozin, Sebastian},
  booktitle={International conference on machine learning},
  pages={3481--3490},
  year={2018},
  organization={PMLR}
}

@article{hacohen2024ltxvideo,
  title={Ltx-video: Realtime video latent diffusion},
  author={HaCohen, Yoav and Chiprut, Nisan and Brazowski, Benny and Shalem, Daniel and Moshe, Dudu and Richardson, Eitan and Levin, Eran and Shiran, Guy and Zabari, Nir and Gordon, Ori and others},
  journal={arXiv preprint arXiv:2501.00103},
  year={2024}
}

@inproceedings{nan2025openvid,
  title={Openvid-1m: A large-scale high-quality dataset for text-to-video generation},
  author={Nan, Kepan and Xie, Rui and Zhou, Penghao and Fan, Tiehan and Yang, Zhenheng and Chen, Zhijie and Li, Xiang and Yang, Jian and Tai, Ying},
  booktitle={International conference on learning representations},
  volume={2025},
  pages={1045--1064},
  year={2025}
}

@inproceedings{chen20254dnex,
title={4{DN}eX: Feed-Forward 4D Generative Modeling Made Easy},
author={Zhaoxi Chen and Tianqi Liu and Long Zhuo and Jiawei Ren and Zeng Tao and He Zhu and Fangzhou Hong and Liang Pan and Ziwei Liu},
booktitle={1st Workshop on Reliable and Interactive World Model in Computer Vision Non Archival},
year={2025}
}

@article{ghosh2023geneval,
  title={Geneval: An object-focused framework for evaluating text-to-image alignment},
  author={Ghosh, Dhruba and Hajishirzi, Hannaneh and Schmidt, Ludwig},
  journal={Advances in Neural Information Processing Systems},
  volume={36},
  pages={52132--52152},
  year={2023}
}

@article{zheng2025vbench2,
  title={Vbench-2.0: Advancing video generation benchmark suite for intrinsic faithfulness},
  author={Zheng, Dian and Huang, Ziqi and Liu, Hongbo and Zou, Kai and He, Yinan and Zhang, Fan and Gu, Lulu and Zhang, Yuanhan and He, Jingwen and Zheng, Wei-Shi and others},
  journal={arXiv preprint arXiv:2503.21755},
  year={2025}
}

@inproceedings{liu2024evalcrafter,
  title={Evalcrafter: Benchmarking and evaluating large video generation models},
  author={Liu, Yaofang and Cun, Xiaodong and Liu, Xuebo and Wang, Xintao and Zhang, Yong and Chen, Haoxin and Liu, Yang and Zeng, Tieyong and Chan, Raymond and Shan, Ying},
  booktitle={IEEE/CVF Conference on Computer Vision and Pattern Recognition (CVPR)},
  pages={22139--22149},
  year={2024},
  organization={IEEE}
}

@article{hu2022lora,
  title={Lora: Low-rank adaptation of large language models},
  author={Hu, Edward J and Shen, Yelong and Wallis, Phillip and Allen-Zhu, Zeyuan and Li, Yuanzhi and Wang, Shean and Wang, Lu and Chen, Weizhu},
  journal={arXiv preprint arXiv:2106.09685},
  year={2021}
}

@article{loshchilov2019decoupled,
  title={Decoupled weight decay regularization},
  author={Loshchilov, Ilya and Hutter, Frank},
  journal={arXiv preprint arXiv:1711.05101},
  year={2017}
}

@article{han2026aesrm,
  title={Aesrm: Improving video aesthetics with expert-level feedback},
  author={Han, Yujin and Wei, Yujie and He, Yefei and Liu, Xinyu and Li, Tianle and Yu, Zichao and Han, Andi and Zhang, Shiwei and Weng, Tingyu and Zou, Difan},
  journal={arXiv preprint arXiv:2604.28078},
  year={2026}
}

@article{xiong2024lvd2m,
  title={Lvd-2m: A long-take video dataset with temporally dense captions},
  author={Xiong, Tianwei and Wang, Yuqing and Zhou, Daquan and Lin, Zhijie and Feng, Jiashi and Liu, Xihui},
  journal={Advances in Neural Information Processing Systems},
  volume={37},
  pages={16623--16644},
  year={2024}
}

@inproceedings{saharia2022photorealistic,
title={Photorealistic Text-to-Image Diffusion Models with Deep Language Understanding},
author={Chitwan Saharia and William Chan and Saurabh Saxena and Lala Li and Jay Whang and Remi Denton and Seyed Kamyar Seyed Ghasemipour and Raphael Gontijo-Lopes and Burcu Karagol Ayan and Tim Salimans and Jonathan Ho and David J. Fleet and Mohammad Norouzi},
booktitle={Advances in Neural Information Processing Systems},
editor={Alice H. Oh and Alekh Agarwal and Danielle Belgrave and Kyunghyun Cho},
year={2022}
}

@article{lu2025onpolicydistillation,
  author = {Kevin Lu and Thinking Machines Lab},
  title = {On-Policy Distillation},
  journal = {Thinking Machines Lab: Connectionism},
  year = {2025},
  note = {https://thinkingmachines.ai/blog/on-policy-distillation},
  doi = {10.64434/tml.20251026},
}

@article{yu2026mismatch,
  title={Mismatch Matters: On-Policy Distillation Beyond Token Agreement},
  author={Yu, Zichao and Yu, Chengzhi and Xu, Shengze and Han, Yujin and Jiang, Bingqing and Wang, Xu and Zou, Difan},
  journal={arXiv preprint arXiv:2608.09836},
  year={2026}
}

@misc{schuhmann2022laion,
  title  = {{LAION-Aesthetics}},
  author = {Schuhmann, Christoph},
  year   = {2022},
  url    = {https://laion.ai/blog/laion-aesthetics/}
}

@article{kirstain2023pick,
  title={Pick-a-pic: An open dataset of user preferences for text-to-image generation},
  author={Kirstain, Yuval and Polyak, Adam and Singer, Uriel and Matiana, Shahbuland and Penna, Joe and Levy, Omer},
  journal={Advances in neural information processing systems},
  volume={36},
  pages={36652--36663},
  year={2023}
}

@inproceedings{wu2023human,
  title={Human preference score: Better aligning text-to-image models with human preference},
  author={Wu, Xiaoshi and Sun, Keqiang and Zhu, Feng and Zhao, Rui and Li, Hongsheng},
  booktitle={2023 IEEE/CVF International Conference on Computer Vision (ICCV)},
  pages={2096--2105},
  year={2023},
  organization={IEEE}
}

@inproceedings{hessel2021clipscore,
  title={Clipscore: A reference-free evaluation metric for image captioning},
  author={Hessel, Jack and Holtzman, Ari and Forbes, Maxwell and Le Bras, Ronan and Choi, Yejin},
  booktitle={Proceedings of the 2021 conference on empirical methods in natural language processing},
  pages={7514--7528},
  year={2021}
}

@article{xu2023imagereward,
  title={Imagereward: Learning and evaluating human preferences for text-to-image generation},
  author={Xu, Jiazheng and Liu, Xiao and Wu, Yuchen and Tong, Yuxuan and Li, Qinkai and Ding, Ming and Tang, Jie and Dong, Yuxiao},
  journal={Advances in Neural Information Processing Systems},
  volume={36},
  pages={15903--15935},
  year={2023}
}

@article{liu2026flow,
  title={Flow-grpo: Training flow matching models via online rl},
  author={Liu, Jie and Liu, Gongye and Liang, Jiajun and Li, Yangguang and Liu, Jiaheng and Wang, Xintao and Wan, Pengfei and Zhang, Di and Ouyang, Wanli},
  journal={Advances in neural information processing systems},
  volume={38},
  pages={40783--40818},
  year={2026}
}

@article{singer2022makeavideo,
  title={Make-a-video: Text-to-video generation without text-video data},
  author={Singer, Uriel and Polyak, Adam and Hayes, Thomas and Yin, Xi and An, Jie and Zhang, Songyang and Hu, Qiyuan and Yang, Harry and Ashual, Oron and Gafni, Oran and others},
  journal={arXiv preprint arXiv:2209.14792},
  year={2022}
}

@article{yin2024improved,
  title={Improved distribution matching distillation for fast image synthesis},
  author={Yin, Tianwei and Gharbi, Micha{\"e}l and Park, Taesung and Zhang, Richard and Shechtman, Eli and Durand, Fredo and Freeman, William T},
  journal={Advances in neural information processing systems},
  volume={37},
  pages={47455--47487},
  year={2024}
}

@inproceedings{yuan2024instructvideo,
  title={Instructvideo: Instructing video diffusion models with human feedback},
  author={Yuan, Hangjie and Zhang, Shiwei and Wang, Xiang and Wei, Yujie and Feng, Tao and Pan, Yining and Zhang, Yingya and Liu, Ziwei and Albanie, Samuel and Ni, Dong},
  booktitle={2024 IEEE/CVF Conference on Computer Vision and Pattern Recognition (CVPR)},
  pages={6463--6474},
  year={2024},
  organization={IEEE}
}

@article{prabhudesai2407video,
  title={Video diffusion alignment via reward gradients (2024)},
  author={Prabhudesai, Mihir and Mendonca, Russell and Qin, Zheyang and Fragkiadaki, Katerina and Pathak, Deepak},
  journal={arXiv preprint arXiv:2407.08737}
}
\bibliographystyle{iclr2027_conference}

\clearpage
\crefalias{section}{appendix}
\crefalias{subsection}{subappendix}
\crefalias{subsubsection}{subsubappendix}
\appendix
\begingroup
\setcounter{tocdepth}{1}
\tableofcontents
\endgroup
\clearpage
\paragraph{Roadmap.} The appendix is organized as follows:
\begin{itemize}[leftmargin=*, itemsep=2pt]
    \item \Cref{app:derivations} provides the detailed derivations behind the proposed method: the closed-form OPD loss from rectified flow, the connector velocity mapping, the anchored image and motion targets together with their correction bounds, and the gradients used for joint optimization.
    \item \Cref{app:implementation_details} gives implementation details: model configurations, training hyperparameters, and evaluation protocols.
    \item \Cref{app:additional_analyses} presents additional analyses, including the motion-reward validation.
\end{itemize}

\section{Detailed derivations}
\label{app:derivations}
\setcounter{secnumdepth}{3}

We use the notation of~\Cref{sec:method} and suppress the conditioning variable $c$ where unambiguous. Local transition calculations use $0<t<1$, $0<h\leq t$, and a diffusion coefficient $\sigma_t>0$ that depends only on time. Density calculations assume the regularity and boundary conditions needed to interchange derivatives and integrals. The SDE argument further assumes well-posedness and a matched initial marginal. Connector parameters are held fixed when differentiating a path with respect to diffusion time.

\subsection{From rectified flow to the closed-form OPD loss}
\label{app:opd}

We derive the Gaussian transition kernels used in Eq.~(\ref{eq:opd_kl}) and evaluate their KL divergence. The rectified-flow interpolation $x_t=(1-t)x_0+t\epsilon$, where $\epsilon\sim\mathcal N(0,I)$ is independent of $x_0$, defines the reference conditional density $p_t(x\mid x_0,c)=\mathcal N(x;(1-t)x_0,t^2I)$. Differentiating its marginal gives
\begin{equation*}
\begin{aligned}
\nabla_x\log p_t(x\mid c)
&=\frac{\int p_{\mathrm{data}}(x_0\mid c)\nabla_x p_t(x\mid x_0,c)\,\mathrm{d}x_0}{p_t(x\mid c)}\\
&=\mathbb E\!\left[-\frac{x-(1-t)x_0}{t^2}\,\middle|\,x_t=x,c\right]\\
&=-\frac{1}{t}\mathbb E[\epsilon\mid x_t=x,c].
\end{aligned}
\end{equation*}
The regression-optimal rectified-flow velocity is $v_t(x,c)=\mathbb E[\epsilon-x_0\mid x_t=x,c]$. Since $\epsilon=x_t+(1-t)(\epsilon-x_0)$, we have
\begin{equation*}
\mathbb E[\epsilon\mid x_t=x,c]=x+(1-t)v_t(x,c).
\end{equation*}
Consequently, the reference score and its regression-optimal velocity satisfy
\begin{equation*}
\nabla_x\log p_t(x\mid c)=-\frac{x+(1-t)v_t(x,c)}{t}.
\end{equation*}
For a learned or constructed velocity $u$, we use the same algebraic form to define the score estimate $\mathcal S_t(x,u):=-[x+(1-t)u]/t$. This estimate need not equal the score of the reference marginal.

For the exact velocity field $v_t$ and reference score $s_t=\nabla\log p_t$, the flow satisfies $\partial_t p_t=-\nabla\cdot(p_t v_t)$. Introducing the increasing reverse clock $\tau=1-t$, consider
\begin{equation*}
\mathrm{d}Y_\tau=\left[-v_t(Y_\tau,c)+\frac{\sigma_t^2}{2}s_t(Y_\tau,c)\right]\mathrm{d}\tau+\sigma_t\,\mathrm{d}B_\tau,\qquad t=1-\tau.
\end{equation*}
For $q_\tau=p_{1-\tau}$, the corresponding Fokker--Planck operator gives
\begin{equation*}
\begin{aligned}
&-\nabla\cdot\left[q_\tau\left(-v_t+\frac{\sigma_t^2}{2}s_t\right)\right]+\frac{\sigma_t^2}{2}\Delta q_\tau\\
&\qquad=\nabla\cdot(q_\tau v_t)-\frac{\sigma_t^2}{2}\nabla\cdot(q_\tau s_t)+\frac{\sigma_t^2}{2}\Delta q_\tau\\
&\qquad=\nabla\cdot(q_\tau v_t)=\partial_\tau q_\tau,
\end{aligned}
\end{equation*}
where the second equality uses $q_\tau s_t=\nabla q_\tau$. Under the stated assumptions, this continuous-time SDE has the reference flow marginals. This result applies to the exact fields $v_t$ and $s_t$.

For distillation, we use the corresponding drift expression to construct local transitions from predicted velocities. In decreasing diffusion time, substituting $\mathcal S_t(x,u)$ gives
\begin{equation}
\begin{aligned}
b_t(x,u)
&=u-\frac{\sigma_t^2}{2}\mathcal S_t(x,u)\\
&=\frac{\sigma_t^2}{2t}x+\chi_tu,
\qquad
\chi_t=1+\frac{\sigma_t^2(1-t)}{2t}.
\end{aligned}
\label{eq:app_reverse_drift}
\end{equation}
Euler--Maruyama discretization over a step $t\to t-h$ yields
\begin{equation*}
x_{t-h}=x_t-hb_t(x_t,u)+\sigma_t\sqrt h\,\xi,\qquad \xi\sim\mathcal N(0,I).
\end{equation*}
At fixed step size $h$, the conditional transition is therefore
\begin{equation}
\pi_u(\cdot\mid x_t,t,c)=\mathcal N(\mu_u,\sigma_t^2hI),
\qquad
\mu_u=\left(1-\frac{h\sigma_t^2}{2t}\right)x_t-h\chi_tu,
\label{eq:app_gaussian_policy}
\end{equation}
where $u$ is evaluated at $(x_t,t,c)$. Evaluating the student and target velocities at the same state cancels the state-dependent term in their means:
\begin{equation*}
\mu_{v_\theta}-\mu_{\bar v}=-h\chi_t(v_\theta-\bar v).
\end{equation*}
Because the two Gaussian kernels share the covariance $\sigma_t^2hI$, their KL divergence is
\begin{equation*}
\begin{aligned}
D_{\mathrm{KL}}\!\left(\pi_{v_\theta}\Vert\pi_{\bar v}\right)
&=\frac{\|\mu_{v_\theta}-\mu_{\bar v}\|_2^2}{2\sigma_t^2h}\\
&=\frac{h\chi_t^2}{2\sigma_t^2}\|v_\theta-\bar v\|_2^2.
\end{aligned}
\end{equation*}
Dividing by the number of video-latent entries $d$ and substituting $\chi_t$ gives
\begin{equation*}
\ell_t(v_\theta,\bar v)
=\frac{1}{d}D_{\mathrm{KL}}\!\left(\pi_{v_\theta}\Vert\pi_{\bar v}\right)
=\frac{h}{2\sigma_t^2}\left(1+\frac{\sigma_t^2(1-t)}{2t}\right)^2\|v_\theta-\bar v\|_{\mathrm{av}}^2,
\end{equation*}
which establishes Eqs.~(\ref{eq:opd_closed_form})--(\ref{eq:opd_weight}). The KL equality is exact for the constructed Gaussian kernels, while the Euler step approximates the continuous-time dynamics. The OPD objective in Eq.~(\ref{eq:opd_kl}) evaluates this local loss at states visited by the current student; it does not assume that $p_\theta(\cdot\mid t,c)$ equals the reference marginal $p_t$.

\subsection{Mapping image guidance to video latents}
\label{app:score_pullback}

\subsubsection{State and velocity mapping}

Consider a differentiable rectified-flow path $x(t)$ and a connector $F_\psi$ whose parameters remain fixed along that path. For $z(t)=F_\psi(x(t))$, differentiability gives
\begin{equation*}
F_\psi(x(t+\Delta t))-F_\psi(x(t))
=J_{F_\psi}(x(t))\dot x(t)\Delta t+o(|\Delta t|).
\end{equation*}
Dividing by $\Delta t$ and taking the limit yields
\begin{equation*}
\dot z(t)=J_{F_\psi}(x(t))\dot x(t).
\end{equation*}
Setting $\dot x(t)=v_\theta(x(t),t,c)$ gives the mapped velocity in Eq.~(\ref{eq:connector_mapping}). The connector may change between training iterations but is fixed within this local diffusion-time derivative. This chain-rule mapping requires no inverse of $F_\psi$ and does not imply exact transport of stochastic transition distributions.

For a bias-free linear connector $F_\psi(x)=A_\psi x$, the Jacobian is constant and $J_{F_\psi}(x)v=A_\psi v$. Fixed linear slice selection and spatial resizing can be included in $A_\psi$, so the complete operator maps both states and velocities.

\subsubsection{Velocity-to-score conversion and score pullback}

The image-expert predictions $u_k(z_t,t,c)$ are velocities in image latent coordinates. Under the rectified-flow convention, we convert each prediction into the image-space score estimate
\begin{equation*}
s_k^{\mathrm I}
=-\frac{z_t+(1-t)u_k(z_t,t,c)}{t}.
\end{equation*}
As in the video-space calculation, this expression equals an image reference marginal's score when $u_k$ is its regression-optimal velocity. For learned experts, it is the score estimate used to construct guidance.

To derive the pullback, first consider exact image scores $s_k^{\mathrm I}(z,t,c)=\nabla_z\log p_{k,t}^{\mathrm I}(z\mid c)$. With the prompt-dependent weights $\beta_k(c)$ fixed, define $\Phi_t(x)=\sum_{k=1}^{K}\beta_k(c)\log p_{k,t}^{\mathrm I}(F_\psi(x)\mid c)$. For any direction $\delta x$,
\begin{equation*}
\begin{aligned}
\left.\frac{\mathrm d}{\mathrm da}\Phi_t(x+a\delta x)\right|_{a=0}
&=\sum_{k=1}^{K}\beta_k(c)(s_k^{\mathrm I})^\top J_{F_\psi}(x)\delta x\\
&=\left[J_{F_\psi}(x)^\top\sum_{k=1}^{K}\beta_k(c)s_k^{\mathrm I}\right]^\top\delta x.
\end{aligned}
\end{equation*}
Because this holds for every $\delta x$, it gives $\nabla_x\Phi_t(x)=J_{F_\psi}(x)^\top\sum_k\beta_k(c)s_k^{\mathrm I}$, as in Eq.~(\ref{eq:score_pullback}). The vector--Jacobian product uses the scores evaluated at the queried state and does not require derivatives through the image experts. It also requires neither an inverse connector nor a change-of-variables determinant. The potential $\Phi_t$ need not define a normalized density over video latents. For learned score estimates, the same pullback defines the local image-expert direction used by the method.

\subsection{Connector alignment and nonlinear training}
\label{app:connector_geometry}

The connector objective in Eq.~(\ref{eq:connector_objective}) combines paired-latent content alignment with velocity alignment on student-visited states. In the content term, $\bar x_0$ and $\bar z_0$ are fixed encodings of the same visual content under the video and image VAEs. In the velocity term, the sampled video state $x_t$ and student velocity $v_\theta(x_t,t,c)$ are fixed inputs, while each expert prediction is a detached target. Thus, the expert's input $z_t=F_\psi(x_t)$ may change as $\psi$ changes, but gradients are not propagated through the expert prediction $u_k(z_t,t,c)$.

For a nonlinear connector, velocity alignment differentiates the Jacobian--vector product $J_{F_\psi}(x_t)v^{\mathrm{det}}$, where $v^{\mathrm{det}}=\operatorname{sg}[v_\theta(x_t,t,c)]$. Assuming the required mixed derivatives exist, its derivative for output coordinate $j$ and parameter coordinate $p$ is
\begin{equation*}
\frac{\partial[J_{F_\psi}(x_t)v^{\mathrm{det}}]_j}{\partial\psi_p}
=\sum_i
\frac{\partial^2F_{\psi,j}(x_t)}
{\partial\psi_p\,\partial x_i}
v^{\mathrm{det}}_i.
\end{equation*}
These mixed input--parameter derivatives allow the velocity term to train a nonlinear connector. For the default linear connector $F_\psi(x)=A_\psi x$, the corresponding term reduces to $A_\psi v^{\mathrm{det}}$. In either case, the connector's velocity term is an image-space mean-squared error; it does not use the reverse-SDE weight $w(t,h)$. The full parameter gradient is given in~\Cref{app:optimization}.

\subsection{Anchored image and motion targets}

\subsubsection{Image-target conversion and correction bound}
\label{app:anchored_target}

Let $s_{\mathrm A}=\mathcal S_t(x_t,v_{\mathrm A})$. For a provisional nonnegative scale $a_t$, adding image guidance $g_{\mathrm I}$ to the anchor score and converting back to velocity gives
\begin{equation*}
\begin{aligned}
\frac{-t(s_{\mathrm A}+a_tg_{\mathrm I})-x_t}{1-t}
&=\frac{-ts_{\mathrm A}-x_t}{1-t}-\frac{t}{1-t}a_tg_{\mathrm I}\\
&=v_{\mathrm A}-\frac{t}{1-t}a_tg_{\mathrm I}.
\end{aligned}
\end{equation*}
This establishes the sign and time factor of the image-guidance correction in Eq.~(\ref{eq:processed_image_target}) before applying $\mathcal C_{\mathrm I}$.

To set its strength, let $M$ be a binary mask selecting video-latent coordinates covered by image supervision. For one sample, define the stabilized RMS
\begin{equation}
r_M(q)=\sqrt{\frac{\|M\odot q\|_2^2}{\|M\|_1+\varepsilon}},
\qquad \varepsilon>0.
\label{eq:app_rms_def}
\end{equation}
Let $\kappa_{\mathrm I}\geq0$ be the image-score RMS multiplier and $\rho_{\mathrm I}^{(0)}\geq0$ the initial velocity-correction cap relative to the anchor. We normalize the image contribution and rescale its induced velocity correction if necessary:
\begin{equation*}
\begin{aligned}
a_t
&=\kappa_{\mathrm I}\frac{r_M(s_{\mathrm A})}{r_M(g_{\mathrm I})+\varepsilon},\\
\delta v_{\mathrm I}
&=-\frac{t}{1-t}a_tg_{\mathrm I},\\
c_{\mathrm I}
&=\min\!\left(1,\frac{\rho_{\mathrm I}^{(0)}r_M(v_{\mathrm A})}{r_M(\delta v_{\mathrm I})+\varepsilon}\right).
\end{aligned}
\end{equation*}
By homogeneity, $r_M(a_tg_{\mathrm I})\leq\kappa_{\mathrm I}r_M(s_{\mathrm A})$. Setting $\alpha_t=c_{\mathrm I}a_t$ gives the initial image-enhanced proposal $v_{\mathrm I}^{(0)}=v_{\mathrm A}+c_{\mathrm I}\delta v_{\mathrm I}=v_{\mathrm A}-[t/(1-t)]\alpha_tg_{\mathrm I}$. Its correction obeys
\begin{equation}
\begin{aligned}
r_M(v_{\mathrm I}^{(0)}-v_{\mathrm A})
&=c_{\mathrm I}r_M(\delta v_{\mathrm I})\\
&\leq\rho_{\mathrm I}^{(0)}r_M(v_{\mathrm A}).
\end{aligned}
\label{eq:initial_image_bound}
\end{equation}
This bound applies to the initial proposal. The processed target $\widetilde v_{\mathrm I}=\mathcal C_{\mathrm I}(v_{\mathrm A},v_{\mathrm I}^{(0)})$ additionally applies temporal processing and magnitude control. A bound on $\widetilde v_{\mathrm I}$ requires the corresponding properties of $\mathcal C_{\mathrm I}$; it does not follow from Eq.~(\ref{eq:initial_image_bound}) alone.

\paragraph{Temporal correction constraints.}
The image and motion branches use the same correction operator with different settings. For a base velocity $b$ and proposed velocity $p$, define $\Delta=p-b$ and its temporal mean $\mu_\Delta=T^{-1}\sum_{\tau=1}^{T}\Delta_\tau$, broadcast across temporal slices. We separate the shared component from the temporal residual and filter the latter:
\begin{equation}
e=\Delta-\mu_\Delta,
\qquad
\widehat\Delta=\mu_\Delta+\eta\mathcal H e,
\label{eq:temporal_correction_filter}
\end{equation}
where $\eta$ controls the residual contribution. The three-point temporal filter is
\begin{equation*}
[\mathcal H e]_\tau
=\frac14e_{\tau-1}+\frac12e_\tau+\frac14e_{\tau+1},
\qquad e_0=e_1,\quad e_{T+1}=e_T.
\end{equation*}

The filtered correction is scaled and added to the base:
\begin{equation}
\mathcal C_{\eta,\boldsymbol\rho}(b,p)
=b+s_b(\widehat\Delta)\widehat\Delta,
\qquad 0\leq s_b(\widehat\Delta)\leq1,
\label{eq:temporal_correction_operator}
\end{equation}
where $\boldsymbol\rho=(\rho_g,\rho_f,\rho_d)$ specifies global, per-slice, and temporal-difference bounds. For one sample, let $r$ denote RMS over all latent entries and $r_\tau$ RMS over channels and spatial entries at temporal slice $\tau$. Define the adjacent-slice difference $[\mathrm D b]_\tau=b_{\tau+1}-b_\tau$. The per-slice and temporal-difference reference levels are
\begin{equation*}
\begin{aligned}
q_{b,\tau}
&=\max\!\left(r_\tau(b),\xi r(b)\right),\\
q_{d,\tau}
&=\max\!\left(r_\tau(\mathrm D b),\xi r(b)\right),
\end{aligned}
\end{equation*}
where $\xi$ supplies a floor when a local reference is small. The scaling factors are
\begin{equation*}
\begin{aligned}
s_g
&=\frac{\rho_g r(b)}{r(\widehat\Delta)+\varepsilon},\\
s_f
&=\min_\tau\frac{\rho_f q_{b,\tau}}{r_\tau(\widehat\Delta)+\varepsilon},\\
s_d
&=\min_{\tau<T}\frac{\rho_d q_{d,\tau}}{r_\tau(\mathrm D\widehat\Delta)+\varepsilon},\\
s_b(\widehat\Delta)
&=\min(1,s_g,s_f,s_d).
\end{aligned}
\end{equation*}
One scalar is applied to every entry of each sample. Writing $\Delta_{\mathrm{safe}}=s_b(\widehat\Delta)\widehat\Delta$, linearity of $\mathrm D$ and homogeneity of RMS give
\begin{equation*}
\begin{aligned}
r(\Delta_{\mathrm{safe}})
&\leq\rho_g r(b),\\
r_\tau(\Delta_{\mathrm{safe}})
&\leq\rho_f q_{b,\tau},\\
r_\tau(\mathrm D\Delta_{\mathrm{safe}})
&\leq\rho_d q_{d,\tau}.
\end{aligned}
\end{equation*}
The operators in the main text are $\mathcal C_{\mathrm I}=\mathcal C_{\eta_{\mathrm I},\boldsymbol\rho_{\mathrm I}}$ and $\mathcal C_{\mathrm M}=\mathcal C_{\eta_{\mathrm M},\boldsymbol\rho_{\mathrm M}}$. Their parameter values are reported in Appendix~\ref{app:implementation_details}.

\subsubsection{Motion-refinement sign and correction bound}
\label{app:motion}

Motion refinement evaluates the clean latent predicted by $\widetilde v_{\mathrm I}$. Let $b_{\mathrm I}=\operatorname{sg}[\widetilde v_{\mathrm I}]$ and define $f(x_0)=R(D^{\mathrm V}(x_0))$. Holding $b_{\mathrm I}$ fixed gives
\begin{equation*}
\begin{aligned}
\widehat x_0&=x_t-tb_{\mathrm I},\\
g_{\mathrm M}
&=\left.\nabla_x f(x-tb_{\mathrm I})\right|_{x=x_t}
=\nabla f(\widehat x_0).
\end{aligned}
\end{equation*}
If $y=D^{\mathrm V}(\widehat x_0)$, $m=\mathcal O(y)$ is the optical flow computed by the frozen estimator, and $R(y)=D_{\mathrm M}(\mathcal O(y))$, the chain rule gives
\begin{equation*}
g_{\mathrm M}
=J_{D^{\mathrm V}}(\widehat x_0)^\top
J_{\mathcal O}(y)^\top
\nabla_mD_{\mathrm M}(m).
\end{equation*}
The decoder, flow estimator, and discriminator remain frozen but retain these input derivatives.

Define the per-sample RMS $r(q)=\sqrt{\|q\|_2^2/d^{\mathrm V}}$ and let $s_{\widetilde{\mathrm I}}=\mathcal S_t(x_t,\widetilde v_{\mathrm I})$. Let $\chi_{\mathrm M}\geq0$ be the gradient RMS multiplier, $\kappa_{\mathrm M}\geq0$ the motion-guidance scale, $\kappa_{\max}\geq0$ the time-factor cap, and $\rho_{\mathrm M}^{(0)}\geq0$ the initial velocity-correction cap. We compute
\begin{equation*}
\begin{aligned}
\bar g_{\mathrm M}
&=\chi_{\mathrm M}\frac{r(s_{\widetilde{\mathrm I}})}{r(g_{\mathrm M})+\varepsilon}g_{\mathrm M},\\
\delta v_{\mathrm M}
&=-\kappa_{\mathrm M}\min\!\left(\frac{t}{1-t},\kappa_{\max}\right)\bar g_{\mathrm M},\\
c_{\mathrm M}
&=\min\!\left(1,\frac{\rho_{\mathrm M}^{(0)}r(\widetilde v_{\mathrm I})}{r(\delta v_{\mathrm M})+\varepsilon}\right).
\end{aligned}
\end{equation*}
Collecting the scalar factors, define
\begin{equation*}
\gamma_t
=c_{\mathrm M}\kappa_{\mathrm M}
\min\!\left(\frac{t}{1-t},\kappa_{\max}\right)
\chi_{\mathrm M}
\frac{r(s_{\widetilde{\mathrm I}})}{r(g_{\mathrm M})+\varepsilon}.
\end{equation*}
The initial motion-refined proposal is $v_{\mathrm M}^{(0)}=\widetilde v_{\mathrm I}+c_{\mathrm M}\delta v_{\mathrm M}=\widetilde v_{\mathrm I}-\gamma_tg_{\mathrm M}$. By homogeneity,
\begin{equation}
\begin{aligned}
r(v_{\mathrm M}^{(0)}-\widetilde v_{\mathrm I})
&=c_{\mathrm M}r(\delta v_{\mathrm M})\\
&\leq\rho_{\mathrm M}^{(0)}r(\widetilde v_{\mathrm I}).
\end{aligned}
\label{eq:initial_motion_bound}
\end{equation}
The sign follows from its effect on the predicted clean latent:
\begin{equation*}
x_t-tv_{\mathrm M}^{(0)}=\widehat x_0+t\gamma_tg_{\mathrm M}.
\end{equation*}
For a nonzero $g_{\mathrm M}$ and a positive scalar $a\to0$, differentiability gives
\begin{equation*}
f(\widehat x_0+ag_{\mathrm M})
=f(\widehat x_0)+a\|g_{\mathrm M}\|_2^2+o(a).
\end{equation*}
Thus, a sufficiently small positive correction follows a local reward-ascent direction for the initial clean latent. This does not guarantee reward improvement after a finite correction, temporal processing, target fusion, or a full rollout. The processed target $\widetilde v_{\mathrm M}=\mathcal C_{\mathrm M}(\widetilde v_{\mathrm I},v_{\mathrm M}^{(0)})$ additionally applies temporal and magnitude control. The fused target is detached before student optimization, so the student loss does not differentiate through the reward gradient.

\subsection{Gradient derivation for joint optimization}
\label{app:optimization}

We differentiate Eq.~(\ref{eq:student_objective}) with sampled states, motion-activation indicators, and constructed targets held fixed. The sampling distribution of the current student is not differentiated.

At each state, let $\mathcal T_t$ contain $(\omega_{\mathrm A},v_{\mathrm A})$ and $(\omega_{\mathrm I},\widetilde v_{\mathrm I})$, together with $(\omega_{\mathrm M},\widetilde v_{\mathrm M})$ when $r_t=1$. Then $Z_t=\sum_{(\omega,q)\in\mathcal T_t}\omega$ and $\bar v_t=Z_t^{-1}\sum_{(\omega,q)\in\mathcal T_t}\omega q$. Expanding the squared residuals gives
\begin{equation*}
\sum_{(\omega,q)\in\mathcal T_t}\omega\ell_t(v,q)
=Z_t\ell_t(v,\bar v_t)
+w(t,h)\sum_{(\omega,q)\in\mathcal T_t}\omega\|q-\bar v_t\|_{\mathrm{av}}^2.
\end{equation*}
With the targets fixed, the second term is independent of student parameters. Hence, the fused-target loss has the same student gradient as the weighted sum of the active video-space losses, although their scalar values differ.

Let $V_{\theta,j}=J_\theta v_\theta(x_{t_j},t_j,c)$ and $e_j=v_\theta(x_{t_j},t_j,c)-\operatorname{sg}[\bar v_{t_j}]$. The student gradient is
\begin{equation*}
\nabla_\theta\mathcal L_{\mathrm{student}}
=\frac{2}{N}\sum_{j=1}^{N}
\frac{Z_{t_j}w(t_j,h_j)}{d^{\mathrm V}}
V_{\theta,j}^{\top}e_j.
\end{equation*}
Image expertise enters this gradient through $\bar v_{t_j}$; the student loss has no separate image-space prediction-matching term. Since the fused target is detached, $\nabla_\psi\mathcal L_{\mathrm{student}}=0$.

For connector optimization, write $B=J_{F_\psi}(x_t)$, $v^{\mathrm{det}}=\operatorname{sg}[v_\theta(x_t,t,c)]$, and $u_k^{\mathrm{det}}=\operatorname{sg}[u_k(F_\psi(x_t),t,c)]$. Hold the latent inputs and these student and expert predictions fixed when differentiating with respect to $\psi$. Define $q=F_\psi(\bar x_0)-\bar z_0$ and $e_k^{\mathrm{conn}}=Bv^{\mathrm{det}}-u_k^{\mathrm{det}}$. Differentiating Eq.~(\ref{eq:connector_objective}) gives
\begin{equation*}
\begin{aligned}
\nabla_\psi\mathcal L_{\mathrm{conn}}
={}&\frac{2\lambda_z}{d^{\mathrm I}}
\mathbb E_{\mathrm{pair}}\!\left[
(J_\psi F_\psi(\bar x_0))^\top q
\right]\\
&+\frac{2\lambda_v}{K d^{\mathrm I}}
\sum_{k=1}^{K}\mathbb E_{\mathrm{rollout}}\!\left[
(J_\psi(Bv^{\mathrm{det}}))^\top e_k^{\mathrm{conn}}
\right]
+\lambda_{\mathrm{reg}}\nabla_\psi\mathcal R(\psi).
\end{aligned}
\end{equation*}
The velocity-alignment term uses mean-squared error without the reverse-SDE weight $w(t,h)$. Its parameter derivative for a nonlinear connector is given in~\Cref{app:connector_geometry}. Detached student predictions and latent inputs imply $\nabla_\theta\mathcal L_{\mathrm{conn}}=0$.

For the initialization regularizer, we use
\begin{equation*}
\mathcal R(\psi)
=\frac{1}{|\mathcal P|}\sum_{p\in\mathcal P}\frac{\|p-p^{(0)}\|_2^2}{d_p},
\qquad
\nabla_p\mathcal R
=\frac{2(p-p^{(0)})}{|\mathcal P|d_p},
\end{equation*}
where $\mathcal P$ is the set of connector parameter tensors, $d_p$ counts entries in tensor $p$, and $p^{(0)}$ is its initial value. Each tensor is normalized by its size before averaging.
\section{Implementation Details}
\label{app:implementation_details}

\paragraph{Checkpoints and optimization.}
The student checkpoints are Wan2.1-T2V-1.3B, Wan2.2-TI2V-5B, and LTXV-2B-0.9.6-dev, referred to in the main text as Wan2.1-1.3B, Wan2.2-5B, and LTX-Video-2B. The image experts are SD3.5-Medium models with separate LoRA adapters~\citep{hu2022lora}.
The GenEval, aesthetic, and OCR experts are trained for 200, 1,800, and 1,800 training steps, respectively.
Expert predictions use a classifier-free guidance scale of 4.5.


For MILD post-training, we sample one prompt per iteration from the 10,000-prompt pool drawn from OpenVid-1M~\citep{nan2025openvid} and LVD-2M~\citep{xiong2024lvd2m}. The current student generates
one on-policy trajectory for an 81-frame video at $480\times832$ resolution, giving a global batch size of one video. We sample four states from this trajectory using the trajectory-index ranges
in~\Cref{tab:training_settings}. The first 50 iterations update only the connector shared by the three image experts for that backbone; the following 200 iterations jointly optimize the student and connector.

During joint training, losses and gradients for all sampled states are evaluated using parameters at the start of the iteration. Gradients are accumulated before either optimizer is stepped; the student optimizer is stepped first, followed by the connector optimizer. Student optimization uses AdamW~\citep{loshchilov2019decoupled} with zero weight decay, a cosine learning-rate schedule, a 5\% learning-rate warm-up ratio, and a minimum learning rate of 10\% of the initial value. Student and connector gradients are clipped at 1.0 and 0.1, respectively. Computation uses BF16, while master parameters and optimizer states are maintained in FP32. We maintain FP32 exponential moving-average weights with decay 0.99. The Wan samplers use a sigma shift of 5.0; LTX-Video uses a guidance rescale of 0.7 and a maximum text length of 256 tokens. Backbone-specific settings are listed in~\Cref{tab:training_settings}.

\paragraph{Connector implementation.}
The default connector applies bilinear resizing followed by $1\times1$, $3\times3$, and $1\times1$ convolutions. All convolutions omit biases and nonlinear activations. The input and hidden channel counts match the student's latent channel count, while the output has 16 channels. We initialize the connector as a near-identity mapping. The latent-alignment, velocity-alignment, and parameter-regularization coefficients in Eq.~(\ref{eq:connector_objective}) are 0.1, 0.01, and 0.001, respectively. Temporal sampling uses the strides in~\Cref{tab:training_settings}, covering all temporal latent slices of the evaluated videos.

\begin{table*}[t]
\centering
\caption{\textbf{Backbone-specific implementation settings.} Trajectory-index ranges specify candidate states for OPD supervision; temporal stride is measured in RGB-frame indices.}
\label{tab:training_settings}
\footnotesize
\setlength{\tabcolsep}{5pt}
\renewcommand{\arraystretch}{1.08}
\begin{tabular}{@{}lccc@{}}
\toprule
\textbf{Setting}
& \textbf{Wan2.1-1.3B}
& \textbf{Wan2.2-5B}
& \textbf{LTX-Video-2B} \\
\midrule
Student denoising steps
& 50 & 50 & 40 \\
OPD trajectory-index range
& 2--47 & 2--47 & 2--37 \\
Student learning rate
& $5\times10^{-7}$ & $3\times10^{-7}$ & $2\times10^{-6}$ \\
Connector learning rate
& $5\times10^{-6}$ & $1\times10^{-5}$ & $2\times10^{-6}$ \\
Connector hidden channels
& 16 & 48 & 128 \\
Image-supervision temporal stride
& 4 & 4 & 8 \\
Supervised latent slices
& 21 & 21 & 11 \\
\midrule
Effective image-score multiplier $\kappa_{\mathrm I}$
& 0.10 & 0.10 & 0.20 \\
Initial image velocity cap $\rho_{\mathrm I}^{(0)}$
& 0.10 & 0.10 & 0.25 \\
Motion guidance scale $\kappa_{\mathrm M}$
& 0.02 & 0.02 & 0.05 \\
Initial motion velocity cap $\rho_{\mathrm M}^{(0)}$
& 0.05 & 0.05 & 0.10 \\
\midrule
Anchor fusion weight $\omega_{\mathrm A}$
& 0.15 & 0.15 & 0.15 \\
Image-target fusion weight $\omega_{\mathrm I}$
& 0.10 & 0.10 & 0.10 \\
Motion-target fusion weight $\omega_{\mathrm M}$
& 0.05 & 0.05 & 0.05 \\
\bottomrule
\end{tabular}
\end{table*}

\paragraph{Temporal correction settings.}
The image and motion branches use the correction operator defined in Appendix~\ref{app:anchored_target}, with different parameters. The image branch uses $\eta_{\mathrm I}=0.25$ and $\boldsymbol\rho_{\mathrm I}=(0.05,0.05,0.10)$, taking $v_{\mathrm A}$ as its base velocity. The motion branch uses $\eta_{\mathrm M}=0.50$ and $\boldsymbol\rho_{\mathrm M}=(0.025,0.025,0.05)$, taking $\widetilde v_{\mathrm I}$ as its base. Both branches use the reference floor $\xi=0.05$. The operator acts across all latent temporal slices without a coverage mask. Its final scalar scales both the shared component and filtered temporal residual; no further filtering or clipping is applied after target fusion.

\paragraph{Motion representation and discriminator.}
For each adjacent-frame flow field $(u_i,v_i)$, we construct $M_i=(u_i,v_i,\sqrt{u_i^2+v_i^2})$ and stack these fields along the temporal dimension. We use the torchvision implementation of RAFT-Small~\citep{teed2020raft} with 12 flow-estimation updates. Motion volumes used for discriminator training are computed offline and cached in FP32. During OPD, flow is recomputed within the differentiable reward computation. The discriminator consists of a 3D convolutional stem and three factorized spatiotemporal residual stages, with spatial convolution followed by temporal convolution in each block. Spatial average pooling produces temporal tokens, which are augmented with temporal positional embeddings and processed by a two-layer Transformer before a scalar-logit output head.

\paragraph{Motion-discriminator data and objective.}
Motion-discriminator training is separate from the 10,000-prompt MILD post-training setup. All three motion discriminators use the same real-video sources and data-construction protocol. The positive collection contains 1,791 videos from OpenVid-1M~\citep{nan2025openvid} and 2,000 videos from 4DNeX-10M~\citep{chen20254dnex}. For each student backbone, we generate one negative video per positive example using its associated caption and corresponding pretrained video model: Wan2.1-1.3B, Wan2.2-5B~\citep{wan2025wan}, or LTX-Video-2B~\citep{hacohen2024ltxvideo}. Generation uses text-only conditioning.

Each positive video also contributes one temporal hard negative, constructed by uniformly sampling frame shuffling, repetition and dropping, temporal jitter, or discontinuous temporal splicing. Reversal and static-video transformations are excluded because they need not imply implausible motion. Each corrupted example remains paired with its source video within the same split. This produces 11,373 entries per discriminator, equally divided among positive videos, model-generated negatives, and temporal hard negatives. \Cref{tab:motion_training_data} reports the training and validation counts.

\begin{table}[t]
\centering
\caption{\textbf{Motion-discriminator training data.} Generated negatives use the corresponding video backbone; counts precede balanced sampling.}
\label{tab:motion_training_data}
\small
\setlength{\tabcolsep}{6pt}
\begin{tabular}{@{}lrr@{}}
\toprule
\textbf{Source} & \textbf{Train} & \textbf{Validation} \\
\midrule
Real videos & 3,412 & 379 \\
Generated videos & 3,412 & 379 \\
Temporal hard negatives & 3,412 & 379 \\
\midrule
Total & 10,236 & 1,137 \\
\bottomrule
\end{tabular}
\end{table}

Let $m^+$ denote real motion volumes and $m^-$ denote generated or temporally corrupted volumes. We use a logistic discrimination objective with R1 regularization~\citep{mescheder2018training}:
\begin{equation*}
\begin{aligned}
\mathcal L_{D_{\mathrm M}}
=\mathbb E_{m^+}
[\operatorname{softplus}(-D_{\mathrm M}(m^+))]
+\mathbb E_{m^-}
[\operatorname{softplus}(D_{\mathrm M}(m^-))]
+\frac{\lambda_{\mathrm{R1}}}{2}
\mathbb E_{m^+}
[\|\nabla_{m^+}D_{\mathrm M}(m^+)\|_2^2].
\end{aligned}
\label{eq:motion_discriminator_objective}
\end{equation*}
We use AdamW with learning rate $2\times10^{-4}$, batch size four, balanced sampling of positive and negative examples, and BF16 computation for 20 epochs. The R1 coefficient is $\lambda_{\mathrm{R1}}=1$, with the penalty evaluated every 16 optimizer steps.

\paragraph{Prompt routing and motion-refinement settings.}
The prompt router detects visible-text requests using text-related keywords and quoted strings. For image-score aggregation, the normalized weights of the GenEval, aesthetic, and OCR experts are $(0.5,0.5,0)$ for ordinary prompts and $(0.4,0.4,0.2)$ for text-related prompts. These weights apply only to score aggregation. Connector training uses paired clean-latent alignment and equally weighted velocity-alignment targets from all three experts for every prompt, including when the OCR score-pullback weight is zero.

The effective image-score multiplier $\kappa_{\mathrm I}$ is the minimum of the configured image-score scale and score-ratio ceiling. It is 0.10 for the Wan backbones and 0.20 for LTX-Video-2B. Motion refinement uses gradient RMS multiplier $\chi_{\mathrm M}=1$ and time-factor cap $\kappa_{\max}=4$. Backbone-specific motion-guidance scales $\kappa_{\mathrm M}$ and initial velocity caps $\rho_{\mathrm M}^{(0)}$ are listed in~\Cref{tab:training_settings}. These coefficients are distinct from the temporal residual scales $\eta_{\mathrm I}$ and $\eta_{\mathrm M}$.

During student updates, at most one of the four sampled states receives motion refinement, provided its diffusion time satisfies $t\in[0.2,0.6]$. If no state is eligible, motion refinement is skipped for that iteration. The fusion weights $(\omega_{\mathrm A},\omega_{\mathrm I},\omega_{\mathrm M})$ are shared across all three backbones, whereas image and motion guidance scales are backbone-specific, as shown in~\Cref{tab:training_settings}. The active fusion weights sum to $Z_t=0.25$ without motion refinement and $Z_t=0.30$ with it. This sum determines both target normalization and the outer student-loss coefficient.

\section{Additional Analyses}
\label{app:additional_analyses}

\subsection{Motion-Reward Validation}
\label{app:motion_validation}

We evaluate the learned motion discriminators through three
complementary metrics.
AUROC measures their ability to distinguish motion volumes
from real videos from those of generated or temporally
corrupted videos.
Pairwise accuracy measures how often a real motion volume
receives a higher reward than its temporally corrupted
counterpart.
Reward--magnitude correlation measures the association between
the discriminator logit and mean optical-flow magnitude.
Together, these metrics assess discrimination performance,
sensitivity to temporal corruption, and association with
motion magnitude.

\begin{table}[t]
\centering
\caption{
\textbf{Validation of the motion discriminators.}
Higher AUROC and pairwise accuracy are better.
Reward--magnitude correlation is reported as a signed value.
}
\label{tab:motion_discriminator_validation}
\footnotesize
\setlength{\tabcolsep}{5pt}
\renewcommand{\arraystretch}{1.05}

\begin{tabular}{@{}lccc@{}}
\toprule
\textbf{Student backbone}
& \textbf{AUROC}
& \textbf{\shortstack{Pairwise\\accuracy}}
& \textbf{\shortstack{Reward--magnitude\\correlation}} \\
\midrule
Wan2.1-1.3B
& 0.9552 & 0.9472 & $-0.0817$ \\
Wan2.2-5B
& 0.9288 & 0.9736 & $-0.1078$ \\
LTX-Video-2B
& 0.8925 & 0.9538 & $-0.0521$ \\
\bottomrule
\end{tabular}
\end{table}

As shown in~\Cref{tab:motion_discriminator_validation},
all three discriminators distinguish real motion from the
evaluated negative samples and consistently rank original
motion above its temporally corrupted counterpart.
The reward--magnitude correlations are small in absolute value,
indicating limited linear association with motion magnitude
on these validation samples.
These findings support sensitivity to the evaluated temporal
corruptions, without establishing general motion understanding
or independence from motion magnitude.

Under this shared data protocol, positive and generated videos
are partitioned independently, while each temporal corruption
remains in the same split as its source video.
The resulting split is disjoint at the video-file level,
but caption-matched positive and generated videos can appear
in different splits.
Frame rates are also not standardized before flow extraction.
The reported results therefore characterize performance under
this validation protocol and do not establish generalization
to unseen caption groups or robustness to frame-rate differences.

\subsection{Comparison with Aesthetic Video Fine-Tuning}
\label{app:aes_teacher_comparison}

\begin{table*}[t]
\centering
\caption{
\textbf{Comparison with direct aesthetic video fine-tuning.}
Aesthetic SFT denotes video models fine-tuned on aesthetic
video data and evaluated directly, without subsequent OPD.
Bold indicates the better result within each backbone.
Higher is better for all metrics.
}
\label{tab:aes_teacher_comparison}
\footnotesize
\setlength{\tabcolsep}{3pt}
\renewcommand{\arraystretch}{1}

\resizebox{\textwidth}{!}{%
\begin{tabular}{@{}lcccccc@{\hspace{9pt}}ccccc@{}}
\toprule
& \multicolumn{6}{c}{\textbf{VBench-2.0}}
& \multicolumn{5}{c}{\textbf{EvalCrafter}} \\
\cmidrule(lr){2-7}
\cmidrule(l){8-12}
\textbf{Method}
& \textbf{Creativity}
& \textbf{Commonsense}
& \textbf{Controllability}
& \shortstack{\textbf{Human}\\\textbf{Fidelity}}
& \textbf{Physics}
& \textbf{AVG.}
& \shortstack{\textbf{Visual}\\\textbf{Quality}}
& \shortstack{\textbf{Text--Video}\\\textbf{Alignment}}
& \shortstack{\textbf{Motion}\\\textbf{Quality}}
& \shortstack{\textbf{Temporal}\\\textbf{Consistency}}
& \shortstack{\textbf{Final Sum}\\\textbf{Score}} \\
\midrule

\multicolumn{12}{@{}l}{\textit{Wan2.1-1.3B}} \\
Aesthetic SFT
& \textbf{49.84} & 61.13 & 24.94 & \textbf{86.45}
& 44.94 & 53.46
& 66.38 & \textbf{58.90} & 54.18 & 63.34 & 242.80 \\
\rowcolor{gray!10}
\textbf{MILD}
& 47.89 & \textbf{64.56} & \textbf{25.17} & 85.47
& \textbf{46.52} & \textbf{53.92}
& \textbf{66.84} & 57.89 & \textbf{54.58}
& \textbf{63.79} & \textbf{243.10} \\

\midrule
\multicolumn{12}{@{}l}{\textit{Wan2.2-5B}} \\
Aesthetic SFT
& \textbf{50.65} & \textbf{63.13} & \textbf{22.94}
& 83.44 & 46.76 & 53.38
& 63.60 & 57.89 & \textbf{54.45} & \textbf{62.88} & 238.82 \\
\rowcolor{gray!10}
\textbf{MILD}
& 49.50 & 62.84 & 21.80 & \textbf{84.89}
& \textbf{54.29} & \textbf{54.66}
& \textbf{63.76} & \textbf{59.12} & 54.43
& 62.79 & \textbf{240.10} \\

\midrule
\multicolumn{12}{@{}l}{\textit{LTX-Video-2B}} \\
Aesthetic SFT
& 57.01 & 47.99 & 15.80 & \textbf{73.35}
& 48.72 & 48.57
& \textbf{57.57} & 47.87 & 55.16 & 64.39 & \textbf{224.99} \\
\rowcolor{gray!10}
\textbf{MILD}
& \textbf{57.23} & \textbf{48.85} & \textbf{16.78}
& 72.38 & \textbf{50.06} & \textbf{49.06}
& 56.58 & \textbf{48.41} & \textbf{55.30}
& \textbf{64.51} & 224.80 \\

\bottomrule
\end{tabular}%
}
\vspace{-3mm}
\end{table*}

\paragraph{Quantitative comparison.}
We compare MILD with Aesthetic SFT video teacher, which directly fine-tunes each pretrained video backbone on aesthetic video data.
Table~\ref{tab:aes_teacher_comparison} shows that MILD achieves higher VBench-2.0 average scores across all three backbones and higher EvalCrafter aggregate scores on both Wan backbones, while scoring slightly lower on LTX-Video-2B (224.80 vs.\ 224.99).
MILD also achieves higher physics scores across all three backbones, while improvements in other dimensions vary by backbone.
These results support the effectiveness of transferring specialized image expertise compared with direct aesthetic video fine-tuning.

\begin{figure*}[t]
    \centering
    \includegraphics[width=\textwidth]{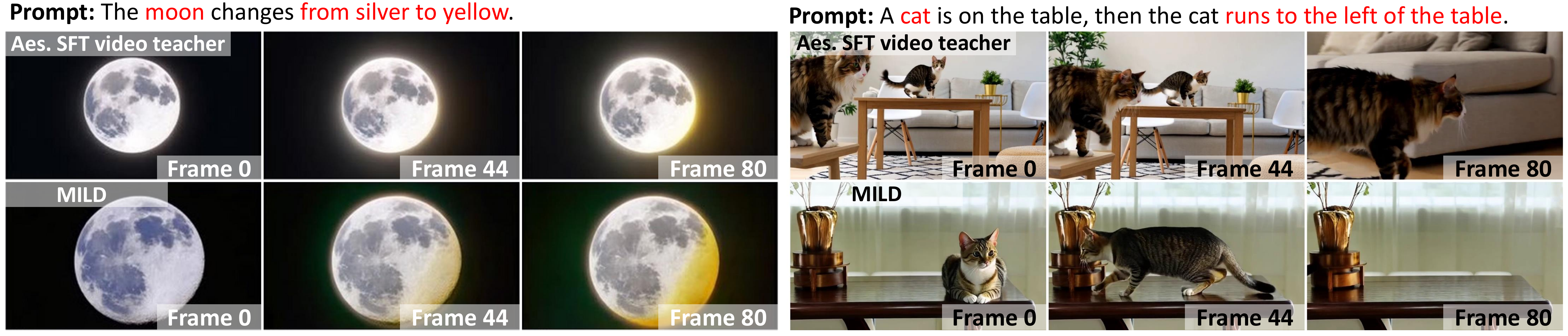}
    \vspace{-3mm}
    \caption{
    \textbf{Qualitative comparison with direct aesthetic
    video fine-tuning on VBench-2.0 using Wan2.2-5B.}
    Frames 0, 44, and 80 illustrate appearance changes
    and directional motion under matched prompts.
    }
    \label{fig:aes_sft_video}
    \vspace{-3mm}
\end{figure*}

\paragraph{Qualitative comparison.}
Figure~\ref{fig:aes_sft_video} compares MILD with direct aesthetic video fine-tuning using Wan2.2-5B on VBench-2.0.
For the moon prompt, MILD shows a clear transition toward yellow while retaining visible lunar texture.
For the cat prompt, MILD depicts a single cat moving left and leaving the frame, whereas Aesthetic SFT introduces an additional cat and does not clearly depict the requested
leftward motion.
These examples complement the quantitative results, illustrating MILD's ability to follow temporal instructions while preserving visual details.

\subsection{Additional Video Case Studies}
\label{app:additional_cases}

\begin{figure*}[t]
    \centering
    \includegraphics[width=\textwidth]{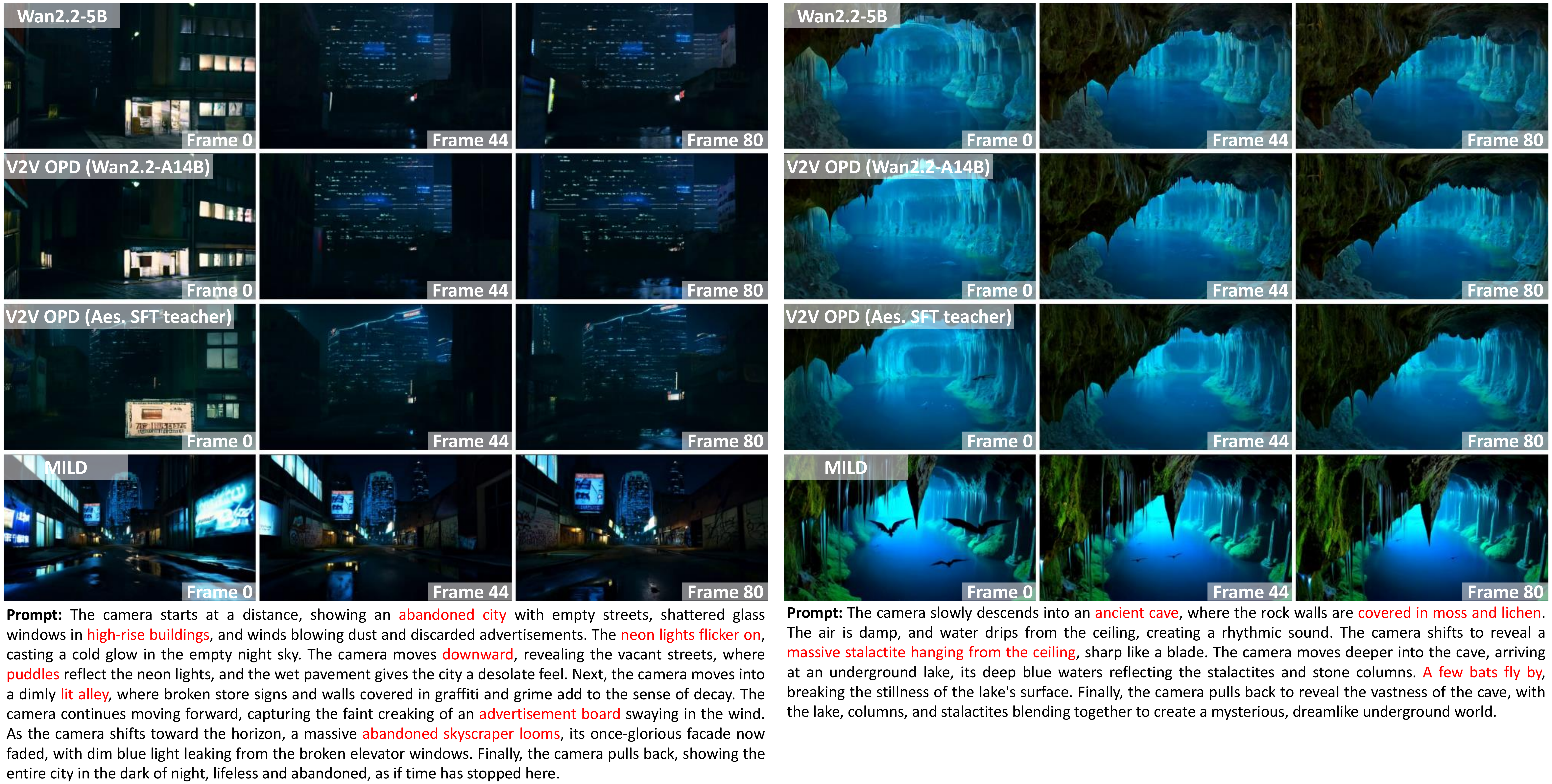}
    \caption{\textbf{Abandoned-city and underground-cave comparisons.} Left: MILD depicts prominent illuminated signs, wall markings, and wet-pavement reflections. Right: MILD renders distinct foreground stalactites, textured rock surfaces, and visible bat silhouettes. Each example shows frames 0, 44, and 80.}
    \label{fig:case_abandoned_cave}
\end{figure*}

We present five additional qualitative comparisons using Wan2.2-5B as the student, complementing the cases in Section~\ref{sec:main_results}. Each example compares the pretrained student, V2V OPD with Wan2.2-A14B, V2V OPD with an aesthetic-SFT teacher, and MILD under the same prompt. Frames 0, 44, and 80 are shown for each generated video. These examples examine scene detail, text legibility, and subject composition.

\paragraph{Scene detail and spatial organization.}
Figure~\ref{fig:case_abandoned_cave} compares the abandoned-city and underground-cave examples. In the abandoned city, MILD depicts prominent illuminated signs, visible wall markings, and reflections on the wet pavement. These details make the foreground street and surrounding buildings easier to distinguish than in the darker baseline outputs. In the cave, MILD renders pronounced foreground stalactites, textured green rock surfaces, and recognizable bat silhouettes above the blue underground lake. The bats are particularly visible in the first sampled frame and appear farther into the cave in the later frames. The foreground rock formations and illuminated background also provide a clear sense of depth.

\begin{figure*}[t]
    \centering
    \includegraphics[width=\textwidth]{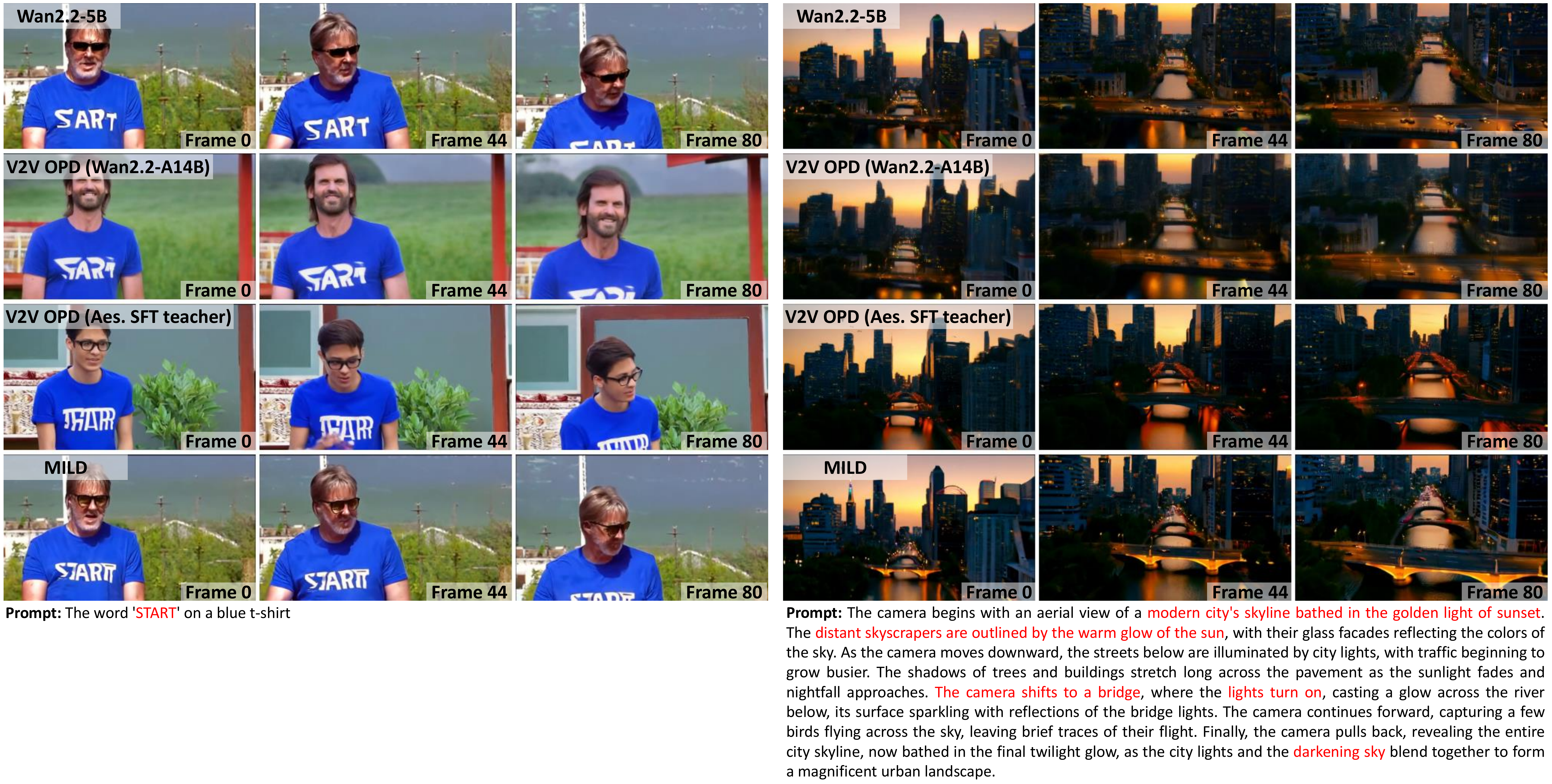}
    \caption{\textbf{Text-rendering and twilight-city comparisons.} Left: MILD renders ``START'' more clearly in the frames where the complete word is visible. Right: MILD depicts distinct architecture, illuminated bridge contours, and water reflections. Each example shows frames 0, 44, and 80.}
    \label{fig:case_text_city}
\end{figure*}

\paragraph{Text rendering and urban scene detail.}
Figure~\ref{fig:case_text_city} presents the blue-T-shirt and twilight-city examples. For the T-shirt prompt, the pretrained student renders ``SART,'' while the V2V OPD outputs contain distorted or ambiguous lettering. MILD renders ``START'' more clearly in the first two sampled frames, where the word is fully visible; the lettering is partially cropped in the final frame. In the twilight-city example, MILD depicts distinct building outlines and illuminated bridges along the central waterway. The bridge contours and their reflections remain readily identifiable across the displayed viewpoints, illustrating detailed architectural content within the generated sequence.

\begin{figure*}[t]
    \centering
    \includegraphics[width=0.60\textwidth]{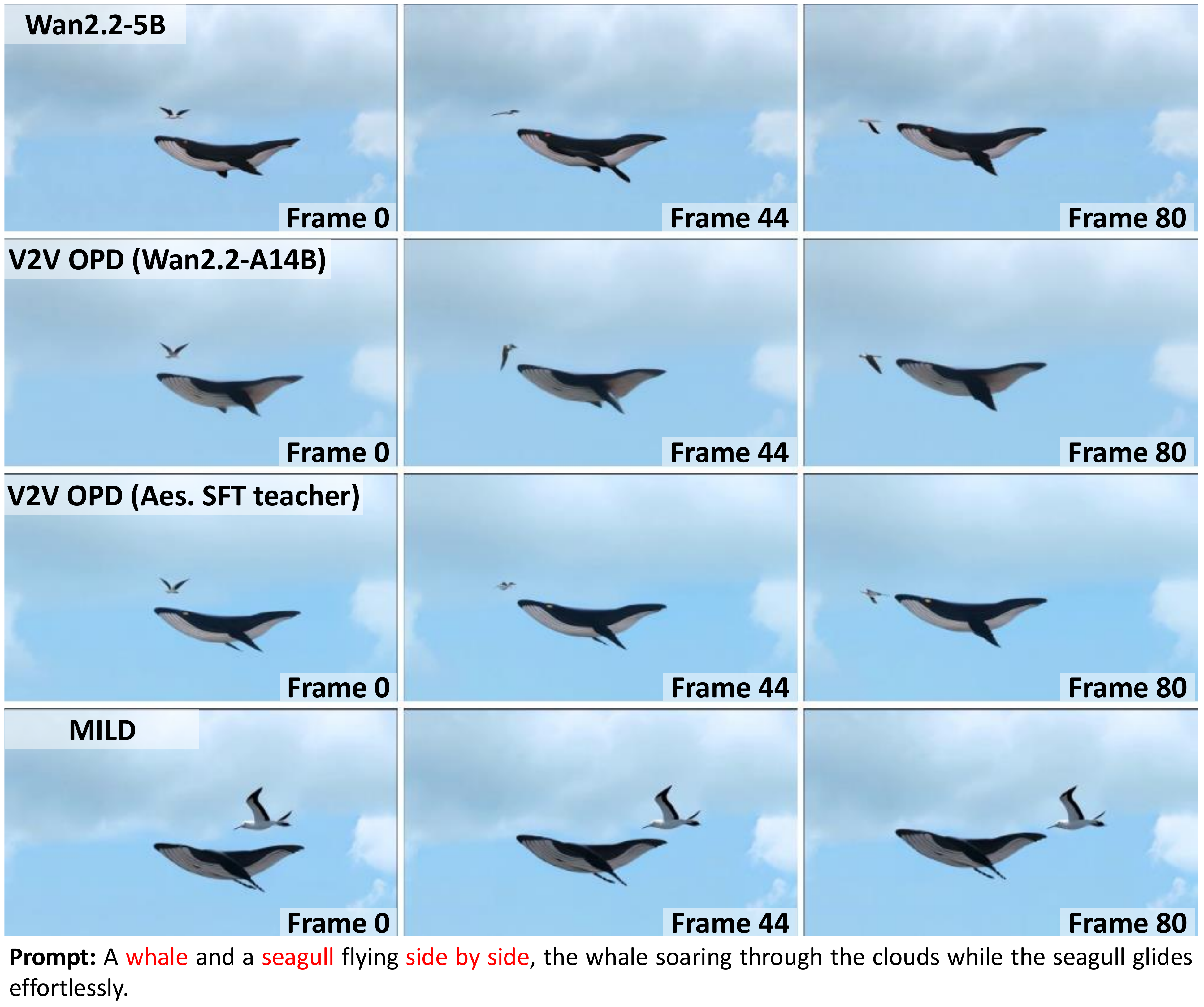}
    \caption{\textbf{Whale-and-seagull composition.} MILD depicts a more prominent seagull alongside the whale, with both subjects clearly identifiable across frames 0, 44, and 80.}
    \label{fig:case_whale_seagull}
\end{figure*}

\paragraph{Subject visibility and composition.}
Figure~\ref{fig:case_whale_seagull} examines the unusual composition of a whale and a seagull flying together. Although both subjects appear in the baseline outputs, their seagulls occupy relatively small image regions. MILD depicts a larger, more prominent seagull with a clearly distinguishable body and wing silhouette alongside the whale. Both subjects remain identifiable in all three sampled frames, making the requested two-subject composition visually explicit.

\subsection{Qualitative Comparison of Aesthetic Teachers}
\label{app:aesthetic_teacher_comparison}

\begin{figure*}[t]
    \centering
    \includegraphics[width=\textwidth]{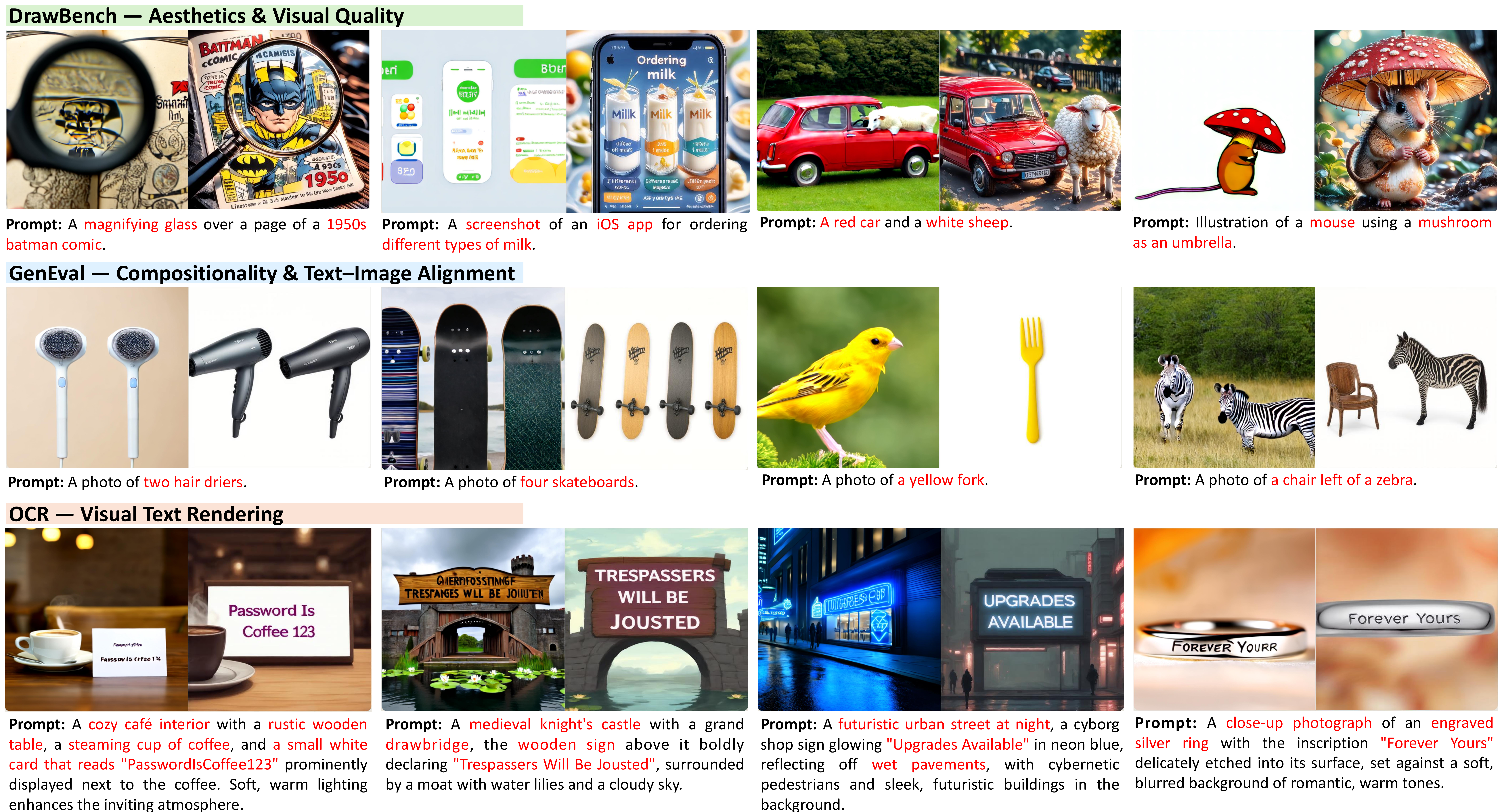}
    \vspace{-3mm}
    \caption{
    \textbf{Qualitative comparison of aesthetic image and video teachers.}
    Each pair shows the Wan2.2-5B video teacher post-trained
    on aesthetic video data (left) and the aesthetic image
    expert (right) under the same prompt.
    Rows show DrawBench, GenEval, and OCR examples, illustrating
    differences in visual detail, compositional accuracy,
    and text rendering available for distillation.
    }
    \label{fig:aesthetic_teacher_comparison}
    \vspace{-3mm}
\end{figure*}

Figure~\ref{fig:aesthetic_teacher_comparison} compares the aesthetic image expert with the Wan2.2-5B video teacher post-trained on aesthetic video data under matched prompts.
These examples examine the visual expertise available for distillation in the Wan2.2-5B student setting, complementing the quantitative teacher comparisons in Figure~\ref{fig:teacher_capabilities}.
On DrawBench prompts, the image expert produces more clearly defined details in the magnifying-glass and comic example and richer texture and lighting in the mouse-and-mushroom scene.
The GenEval cases further reveal differences in compositional accuracy: the image expert depicts four skateboards, a yellow fork, and a chair to the left of a zebra, whereas the video teacher misses the requested count, object identity, or relation.
On OCR prompts, the image expert renders phrases such as ``Trespassers Will Be Jousted,'' ``Upgrades Available,'' and ``Forever Yours'' more legibly and accurately.
However, improved legibility does not always imply exact text reproduction, as illustrated by the altered spacing in the coffee example.
Together, these selected cases illustrate spatial capabilities that image experts can offer for video distillation; they do not assess temporal quality or establish student gains.

\subsection{Qualitative Comparisons of Spatial Capabilities}
\label{app:spatial_qualitative}

\begin{figure*}[t]
    \centering
    \includegraphics[width=\textwidth]{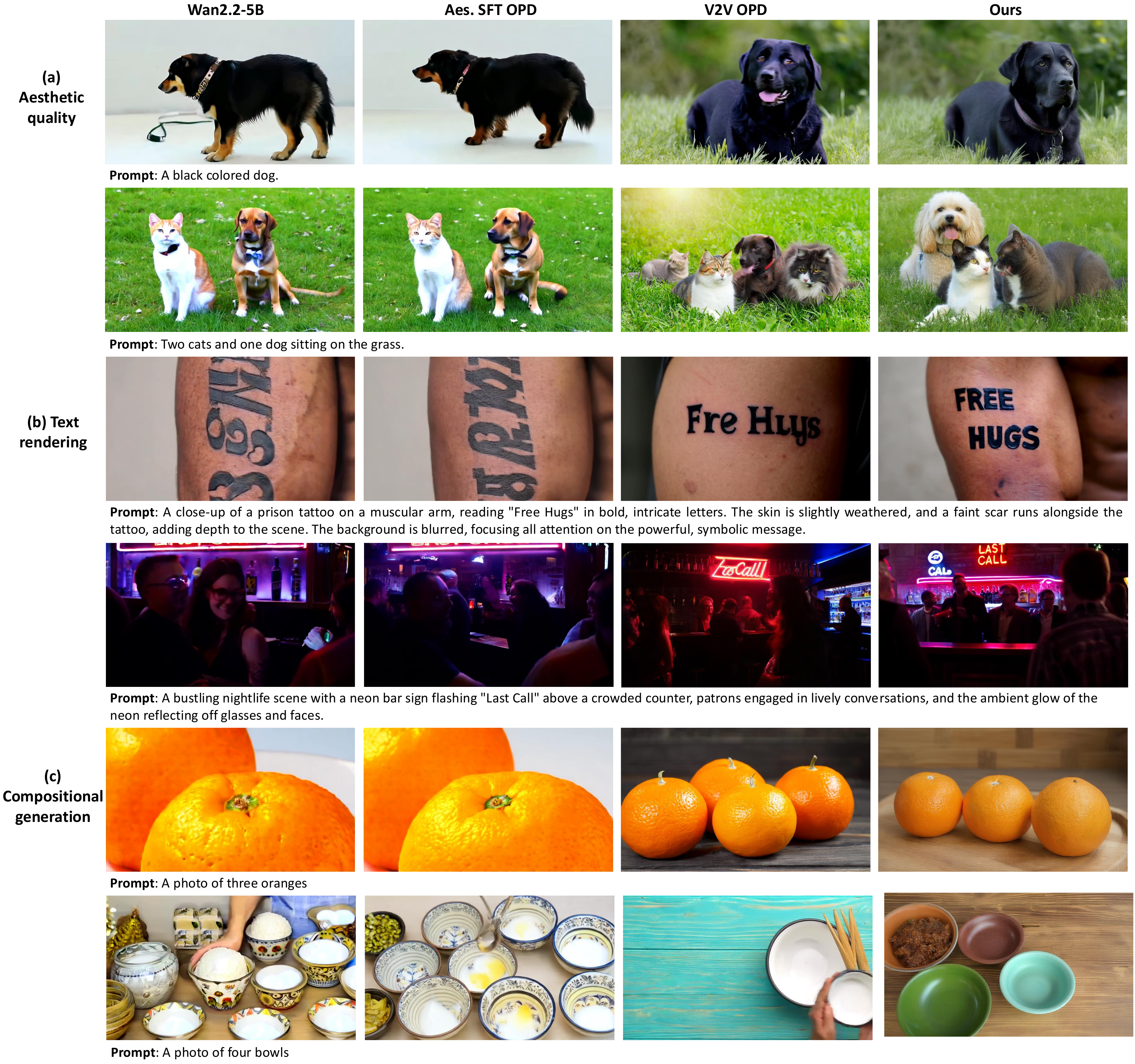}
    \caption{
    \textbf{Spatial capability comparisons using Wan2.2-5B.}
    Columns show the pretrained student, OPD with an
    aesthetic-SFT video teacher, OPD with a larger video
    teacher, and MILD.
    Matched-prompt examples illustrate color-attribute
    alignment, text rendering, and object composition
    in sampled video frames.
    }
    \label{fig:spatial_capability_cases}
\end{figure*}

\begin{figure*}[t]
    \centering
    \includegraphics[width=\textwidth]{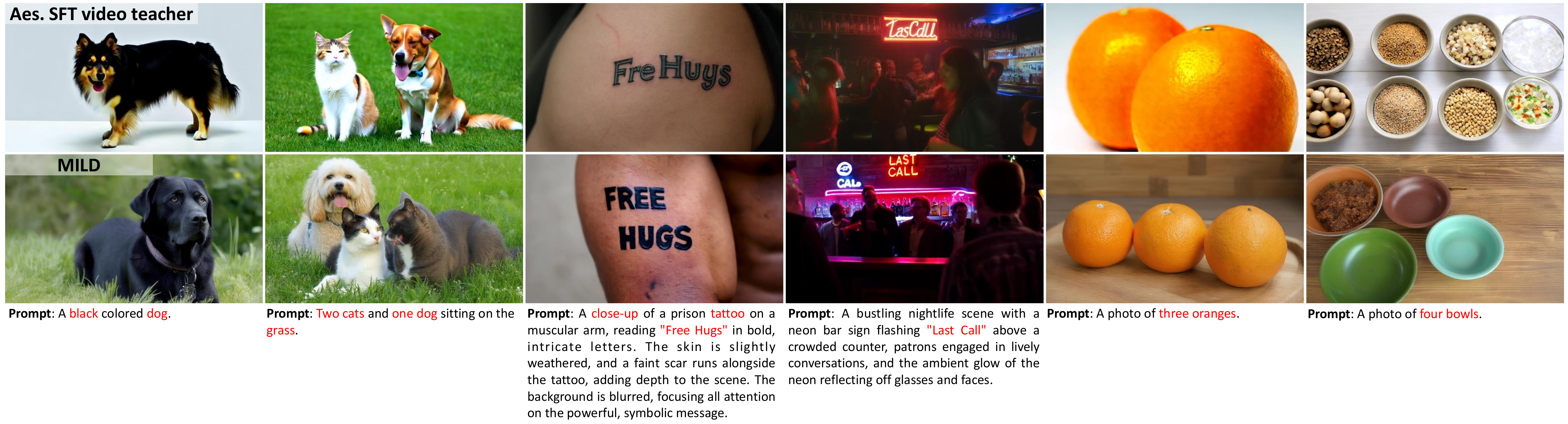}
    \vspace{-3mm}
    \caption{
    \textbf{Spatial capability comparison with direct
    aesthetic video fine-tuning using Wan2.2-5B.}
    Matched-prompt examples illustrate color-attribute
    alignment, text rendering, and object composition
    in sampled video frames.
    }
    \label{fig:aes_sft_spatial}
    \vspace{-3mm}
\end{figure*}

Figure~\ref{fig:spatial_capability_cases} compares MILD with the pretrained Wan2.2-5B student and two V2V OPD baselines under matched prompts, complementing the quantitative evaluation in Section~\ref{sec:spatial_transfer}.

\paragraph{Visual appearance and attribute alignment.}
For the black-dog prompt, MILD depicts a predominantly black coat with visible fur shading and facial details. 
The pretrained student and aesthetic-SFT-teacher OPD produce dogs with substantial tan or white markings, whereas the larger-video-teacher baseline also captures the requested black coat.
This example illustrates MILD's ability to combine subject detail with color-attribute alignment.

\paragraph{Text rendering.}
MILD renders ``FREE HUGS'' clearly on the arm and ``LAST CALL'' legibly on the neon sign.
The pretrained student and aesthetic-SFT-teacher OPD do not clearly reproduce the requested phrases, while the larger-video-teacher baseline produces incomplete or altered text, such as ``Fre Hugs''.
These examples illustrate improved text fidelity on both skin and illuminated signage.

\paragraph{Compositional generation.}
MILD depicts the requested two cats and one dog, three oranges, and four bowls.
The pretrained student and aesthetic-SFT-teacher OPD omit one cat in the animal example.
In the orange example, these two baselines show only two visible oranges, while the larger-video-teacher baseline depicts four.
The bowl examples likewise show count mismatches among the baselines.
Together, these cases illustrate more accurate object counts and category composition with MILD.

\paragraph{Comparison with direct aesthetic video fine-tuning.}
Figure~\ref{fig:aes_sft_spatial} further compares MILD with Aesthetic SFT video teacher, evaluated directly without subsequent OPD.
Using Wan2.2-5B, MILD more closely matches the requested black coat, clearly renders ``FREE HUGS'' and ``LAST CALL'', and satisfies the specified object counts.
Aesthetic SFT exhibits color-attribute, text, and counting errors in these examples.
These comparisons illustrate MILD's advantages in spatial prompt fidelity over direct aesthetic video fine-tuning.

\end{document}